# Topology-aware and highly generalizable deep reinforcement learning for efficient retrieval in multi-deep storage systems

Funing Li[1] · Yuan Tian[2] · Ruben Noortwyck[1] · Jifeng Zhou[1] · Liming Kuang[3] · Robert Schulz[1]

**Abstract**

In modern industrial and logistics environments, the rapid expansion of fast delivery services has heightened the demand for storage systems that combine high efficiency with increased density. Multi-deep autonomous vehicle storage and retrieval systems present a viable solution for achieving greater storage density. However, these systems encounter significant challenges during retrieval operations due to lane blockages. A conventional approach to mitigate this issue involves storing items with homogeneous characteristics in a single lane, but this strategy restricts the flexibility and adaptability of multi-deep storage systems. Building on this background, this work presents a deep reinforcement learning-based framework to optimize the retrieval process in multi-deep storage systems with heterogeneous item configurations. Each item is associated with a specific due date, and objective function is total tardiness minimization. To effectively capture the system's topology, we introduce a graph-based state representation that integrates both item attributes and the local topological structure of the multi-deep warehouse. For processing this representation, we design a novel neural network architecture that combines a Graph Neural Network (GNN) with a Transformer model. The GNN encodes topological and item-specific information into embeddings for all directly accessible items, while the Transformer maps these embeddings into global priority assignments. The Transformer's strong generalization capability further allows our approach to be applied to storage systems with diverse layouts. Extensive numerical experiments, including comparisons with heuristic methods, demonstrate the superiority of the proposed neural network architecture and the effectiveness of the trained agent in optimizing retrieval tardiness.

## Introduction

The recent surge in e-commerce places unprecedented pressure on traditional warehousing and distribution systems. Data from 43 countries show a nearly 60% growth in business-to-consumer e-commerce revenues between 2016 and 2022, hitting USD 27 trillion (United Nations Conference on Trade and Development, 2024). Meanwhile, rising consumer expectations for rapid order fulfillment have compelled many retailers to offer same-day or next-day delivery

✉ Funing Li
funing.li@ift.uni-stuttgart.de

✉ Yuan Tian
tian_yuan4@ctg.com.cn

Ruben Noortwyck
ruben.noortwyck@ift.uni-stuttgart.de

Jifeng Zhou
st181366@stud.uni-stuttgart.de

Liming Kuang
liming.kuang@tum.de

Robert Schulz
robert.schulz@ift.uni-stuttgart.de

[1] Institute of Mechanical Handling and Logistics, University of Stuttgart, Holzgartenstraße 15B, 70174 Stuttgart, Germany

[2] China Yangtze Power Co.,Ltd, No.1 Xiba jianshe Road, Yichang 443000, Hubei, China

[3] Chair for Computer Aided Medical Procedures & Augmented Reality, Technical University of Munich, Boltzmannstraße 3, 85748 Garching, Germany

(Klapp et al., 2020). Reflecting this trend, the global same-day delivery market is projected to grow at a compound annual growth rate of 20.6% from 2025 to 2030 (Grand View Research, 2024).

Consequently, manufacturers and logistics providers face growing challenges in ensuring timely deliveries while scaling operational capacity. To meet these service level requirements while limiting costs, automated warehouses–equipped with intelligent routing and handling strategies–have emerged as key solutions.

In this context, autonomous vehicle storage and retrieval systems (AVS/RS) have gained broad adoption in the warehousing industry owing to their economic advantages, including high operational efficiency and a high degree of automation (Azadeh et al., 2019). Such systems typically consist of multiple tiers. Each tier is served by shuttles that manage the storage and retrieval of items, whereas lifts connect the system to external input/output (I/O) points.

Regarding the lane depth, the AVS/RS can be distinguished into single-deep storage and multi-deep storage (Lerher et al., 2024). As shown in Fig. 1, multi-deep storage offers considerably higher storage density and space efficiency, making it appealing in regions with high land and building costs (Goetschalckx & Donald Ratldff, 1991; Yu & De Koster, 2012). Although multi-deep storage maximizes space utilization, the corresponding retrieval process becomes more complex. This multi-deep configuration requires retrieving items in a last-in-first-out (LIFO) order.

To simplify the retrieval process, several studies assume that each lane contains only a single type of item (Yang et al., 2023; Lupi et al., 2024). However, this assumption fails to fully exploit the high storage density advantage and significantly restricts the applicability of multi-deep storage systems (Accorsi et al., 2017). Moreover, when considering the critical factor of timely delivery, urgent items may be stored at varying depths owing to the inherent randomness of customer orders (Jiang et al., 2021). Furthermore, accounting for product due dates is crucial across industries–ranging from fresh food and newspapers to pharmaceuticals–and requires more efficient and sophisticated management strategies (Zarreh et al., 2024).

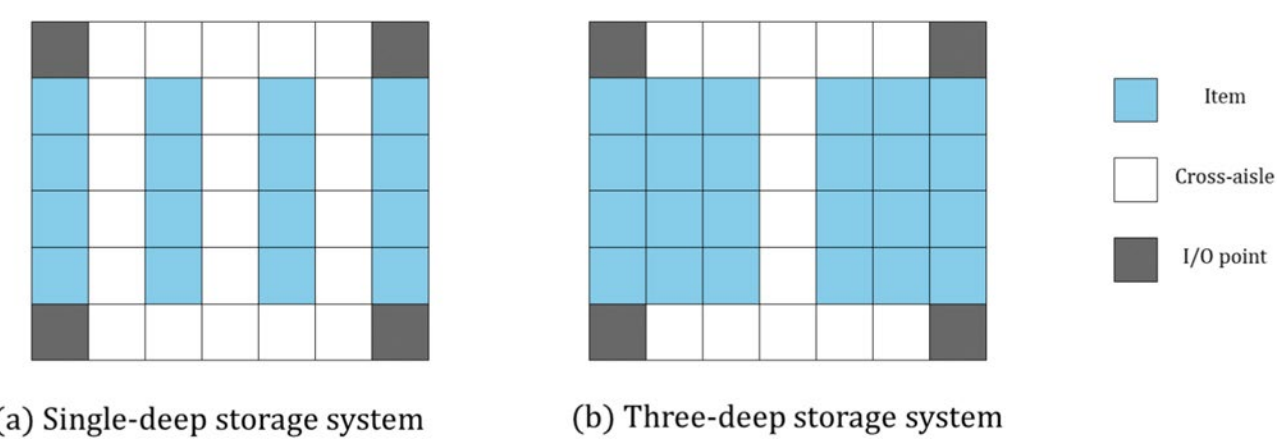

(a) Single-deep storage system (b) Three-deep storage system

**Fig. 1** Example of a single-deep storage system and a three-deep storage system as a representative of multi-deep configurations. The three-deep storage system can store more items within the same warehouse dimension

Unfortunately, most existing studies on multi-deep storage warehouse operations rely heavily on heuristics (Yang et al., 2023; Marolt et al., 2023; Eder, 2020; Lupi et al., 2024; Dong et al., 2021). These heuristics primarily depend on pre-defined rules, which are insufficient to handle variability and unpredictability in modern warehousing scenarios. From a practical perspective, this rigidity often leads to expensive inventory imbalances. In 2023, such imbalances cost companies an estimated USD 1.77 trillion worldwide, with stock-outs accounting for USD 1.2 trillion and overstocks for USD 562 billion (IHL Group, 2023). Consequently, more adaptive decision-making and optimization policies are urgently required to balance accessibility and urgency. This trend is exemplified by forecasts for the global warehouse management system market, which is projected to rise from USD 4 billion in 2024 to USD 8.6 billion by 2029, driven by advances in artificial intelligence and by manufacturers' and logistics practitioners' desire for more responsive operations (MarketsandMarkets, 2024).

With the advancing development of machine learning (ML), deep reinforcement learning (DRL) as a crucial subdomain has yielded promising achievements in various decision-making fields, such as discovering novel matrix multiplication method (Fawzi et al., 2022), mastering strategy video games (Silver et al., 2018; Tian et al., 2023), controlling robotic behaviors (Han et al., 2024), as well as solving challenging planning (Esteso et al., 2023; Tian et al., 2025) and optimization problems (Tian et al., 2020). Online DRL methods employ deep neural networks as agents and train them through interactions with the environment (Sutton and Barto, 2018). The DRL-agent performs actions based on the observed states and receives feedback (rewards) from the environment. Through this interactive process, the agent learns to maximize the rewards, thereby optimizing its decision-making policy.

Motivated by breakthroughs in deep reinforcement learning (DRL), researchers have explored DRL-based approaches to tackle scheduling problems in warehouses (Yan et al., 2022). For instance, Li et al. (2024) develop a DRL-based method to address a machine scheduling problem in the warehouse under the setup constraint. After training with a novel two-stage strategy, the DRL agent can outperform three dispatching rules and two meta-heuristics within a short computation time.

Despite DRL's substantial progress in warehouse optimization, its application to the intricate structural topology of multi-deep storage remains largely unexplored. Moreover, many current solutions have been designed for specific warehouse configurations, limiting their adaptability and real-world practicality. As illustrated in Fig. 2, an item's retrieval path is inherently constrained by the lane design, emphasizing the necessity of incorporating detailed spatial and structural information. Such integration is crucial for developing a versatile DRL agent that is capable of efficiently managing diverse storage layouts and operational scenarios.

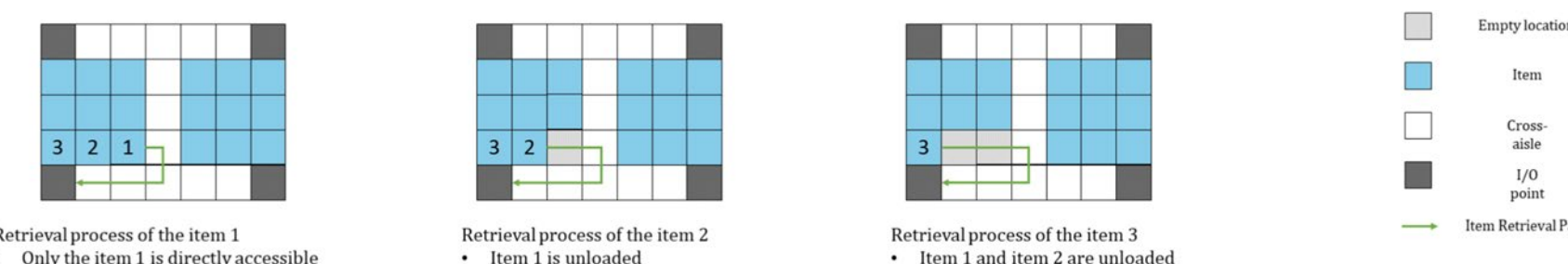

**Fig. 2** Retrieval process of all items in a lane with three-deep. An inner item can only be accessed after all its front items have been retrieved. Items can only move horizontally within the lane. Vertical movement is only allowed at cross-aisles

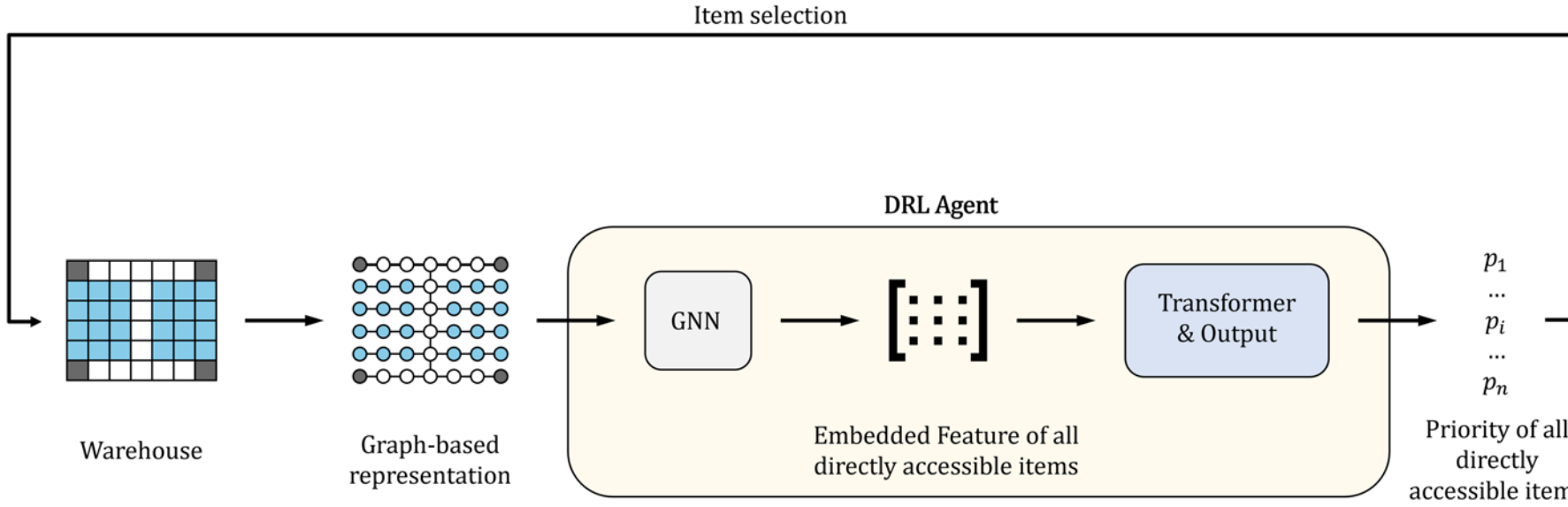

**Fig. 3** Overview of the proposed framework

To address these gaps, we draw inspiration from the robust generalizability of advanced transformer architectures and topological insights offered by graph networks. Specifically, we propose a DRL-based approach for retrieval problem in multi-deep storage systems, to minimize total tardiness. By leveraging detailed map and lane information, our method enables the development of a single agent that can effectively manage both storage systems with various layout configurations and scenarios once trained. We focus on a shuttle-based storage and retrieval system (SBS/RS), a particular type of AVS/RS, and investigate the retrieval process within one tier and a single shuttle. Detailed assumptions are provided in Sect. 3.

The contributions of this work are fourfold:

1. *Heterogeneous-Item Configuration:* We address a more realistic scenario where stored items are heterogeneous, each with its own due date. Unlike the common assumption of identical items per lane, this setup fully leverages the capacity of multi-deep storage systems, thereby offering greater practical applicability and relevance to real-world conditions.
2. *Graph-Based State Representation:* We formulate the warehouse into a graph-based state representation, that can capture both the features of all items and the complex structural relationships between them in a warehouse. Corresponding to this representation, we design a DRL agent integrating GraphSAGE (Hamilton et al., 2017), a specific variant of the graph neural network (GNN). In particular, the GNN model aggregates item features within each lane of the multi-deep storage system. This processing mechanism enables comprehensive and optimal retrieval decisions. Additionally, we define the action space as the set of all accessible items, where the shuttle path is planned by the Dijkstra algorithm (Dijkstra, 1959). This action representation eliminates invalid actions and enhances computational efficiency.
3. *High Scalability:* Building on the aggregated features generated by GraphSAGE, we incorporate a Transformer model (Vaswani et al., 2017) to compute global priorities across all accessible items. This architecture confers high scalability, allowing the approach to adapt seamlessly to warehouses of varying sizes without retraining. By leveraging the combined strengths of GNNs and Transformers, our DRL agent can make comprehensive and optimal retrieval decisions in multi-deep storage environments.
4. *Extensive Numerical Evaluation:* We validate the proposed framework through extensive numerical experiments, including ablation studies and comparisons with the established dispatching rules. The results validate its effectiveness and high generalization capability in minimizing total tardiness.

In summary, this study tackles the retrieval-scheduling problem in multi-deep storage systems where items are heterogeneous and subject to individual due dates. The warehouse is modeled as a graph, and a DRL agent with a hybrid neural network architecture, combining lane-level GraphSAGE aggregation and a Transformer-based priority module, is devised. Extensive numerical experiments demonstrate that the approach scales seamlessly to larger layouts and markedly reduces the total tardiness. A conceptual overview is depicted in Fig. 3.

This paper proceeds as follows. Section 2 is organized into two parts. The first part reviews traditional non-learning-based methods for warehouse management, while the second part examines the utilization of DRL to warehouse scheduling. Section 3 provides the problem description. Section 4 details the proposed approach. Section 5 presents the numerical experiments. Section 6 discusses how the proposed approach could be applied. Section 7 draws conclusions.

## Literature review

In accordance with the topic of this work, the literature review focuses in following two aspects: (1) warehouse management with non-learning-based methods, and (2) warehouse management with learning-based methods, especially DRL-based methods.

### Non-learning-based methods

#### Mathematical programming-based methods

In pursuit of the optimal warehouse management solution, scholars have formulated the problem into a mathematical model and applied optimization techniques to solve it. For instance, Makui et al. (2016) consider an aggregate production planning (APP) problem. It is described by robust optimization model then a Benders decomposition algorithm is leveraged to reduce the computational complexity. In a related study, Aazami & Saidi-Mehrabad (2019) address a robust APP problem under competitive market conditions, where pricing decisions and demand uncertainty are jointly considered through a Stackelberg game framework. This problem is converted into a single-level model, then handled by a Benders decomposition algorithm. Heydari & Aazami (2018) develop a bi-objective mixed-integer nonlinear programming (MINLP) model for a job shop scheduling problem. It is linearized as a mixed-integer linear programming (MILP) model and then tackled by $\epsilon$-constraint method. The computational results illustrate that the method can simultaneously reduce the maximum tardiness as well as completion time. Hu et al. (2024) address a parallel machine scheduling problem. They develop an MILP model and optimally solve small instances (up to 20 jobs) using a commercial solver. Owing to the NP-hard nature, they further develop a neighborhood search-based meta-heuristic for large instances (up to 200 jobs), which achieves better solution quality and faster computation compared to the MILP approach.

#### Meta-heuristic methods

Under similar motivation, the meta-heuristics are broadly utilized for solving scheduling and planning problems (Ezugwu, 2024; Pellerin et al., 2020; Para et al., 2022). Aazami & Saidi-Mehrabad (2021) develop a planning model for fixed-lifetimes products. To address the resulting NP-hard mixed-integer program, they propose a hierarchical heuristic that combines Benders decomposition and a genetic algorithm (GA). Computational experiments on a real-world industrial case demonstrate that the approach efficiently generates high-quality solutions across various problem sizes, effectively balancing seller profits and product freshness. Goli et al. (2018) address a vehicle routing problem with consideration of competitive condition and the customers satisfaction. They propose a meta-heuristic based on cuckoo optimization algorithm. It can exhibit a sufficient solution in a reasonable time. Saeedi Mehrabad et al. (2017) propose a particle swarm algorithm for tackling the location-allocation problem in supply chain. The proposed method jointly minimizes the total cost and maximizes customer satisfaction, and its effectiveness was validated through a comparison with another meta-heuristic and a real-world case study. Azad et al. (2019) investigate an integrated production-inventory-routing planning problem with the consideration of both economical and environmental dimensions. They first mathematically describe the problem in the form of MILP, and then develop a two-stage GA to solve it. Although meta-heuristics are computationally cheaper than the mathematical programming-based approach, they require extensive iterations to yield a proper solution for single instance. This need hinders their applications in environments where instant decision are essential.

#### Rule-based methods

For obtaining real-time solutions, several studies have employed rule-based methods to address the scheduling problem in storage systems. Xiong et al. (2017) consider a job shop scheduling problem. They propose four dispatching rules, which are computationally efficient and exhibit sufficient results in the scenario with loose due date. Yang et al. (2023) address the shuttle scheduling problem in a multi-deep storage system. A two-stage algorithm is proposed, stage one determines the item retrieval sequence according to three rules, whereas stage two calculates the shuttle transfer sequence utilizing dynamic programming. By integrating these rules, the proposed method can yield sufficient solutions within a reasonable time. Dong et al. (2021) investigate retrieval scheduling in an AS/RS containing crane and shuttles and propose a rule-based scheduling method named Lowest-Waiting-Time First (LW). In numerical experiments, LW demonstrates advantages over both the mixed-integer programming (MIP)-based method together with GA. While the MIP approach yields optimal solutions on small-scale instances, its computation time increases exponentially with the problem size. GA achieves better quality than rule-based approaches for small cases, but LW demonstrates the best balance of solution effectiveness and runtime performance when scaled to practical problem sizes.

Nevertheless, the fixed structure of rule-based approaches leads to inferior outcomes in scenarios with altered environmental parameters, such as changed due date range. This lack of adaptability makes them struggling in fluctuating manufacturing contexts. Additionally, crafting or choosing new rule sets for various scenarios demands extensive domain expertise and is labor-intensive. Hence, there is a strong desire for methods capable of delivering real-time solutions in evolving environments.

## DRL-based methods

Unlike mathematical programming-based, meta-heuristic, and rule-based methods, model-free DRL autonomously discovers optimal policies through interaction with environment, without any expert knowledge or handcrafted models (Sutton and Barto, 2018). Once trained, a DRL agent can perform real-time inference to swiftly adapt to evolving conditions, and its trial-and-error learning process endows it with an inherent generalization ability, enabling it to cope with uncertainty, noise, and unforeseen perturbations. Crucially, by optimizing cumulative rewards, DRL not only excels in making efficient short-term decisions but also learns context-appropriate actions that maximize long-term performance, making it exceptionally well-suited for complex, dynamic tasks. For instance, Degrave et al. (2022) employ DRL to solve the tokamak plasmas control problem, which is highly complex and dynamic, featuring a 19-dimensional control space and millisecond-level ($\geq$ 1 kHz) update requirements. The DRL agent trained entirely in simulation could be directly deployed to solve the control problem without any parameter tuning and achieve a significantly low tracking error, demonstrating both real-time performance and strong generalization to novel operating conditions. Meanwhile, because of the integration of deep neural networks, this approach can inherently process high-dimensional data. Vinyals et al. (2019) employ DRL to play a computer game called StarCraft, and the agent could also perform real-time decision making with performance exceeding 99.8% of human players. Notably, the DRL agent processes raw spatial observations, namely seven 64$\times$64 feature planes plus detailed per-unit states, amounting to over 50000 input dimensions, and demonstrating its capability to handle high-dimensional data effectively.

Based on the current success of the RL, many studies have explored its application in addressing various warehousing problems. This highlights the potential of DRL algorithms in optimizing complex decision-making processes and outperforming traditional heuristic approaches. A prominent area of application is storage and/or retrieval optimization. For instance, He et al. (2023) employ a DRL algorithm to solve retrieval problem in a puzzle-based warehouse, in which the storage and retrieval processes rely on escorting items rather than aisles. The proposed approach demonstrated superiority over three heuristic rules across instances of varying scales. For the storage policy, Luo et al. (2021) use a DRL algorithm to optimized the storage process in an SBS/RS of six aisles, single-lane depth, and six tiers, achieving a 6.67% reduction in makespan. Kaynov et al. (2024) propose a DRL-based method for optimizing distribution in a One-Warehouse Multi-Retailer system. Their approach outperformed general-purpose benchmark policies across different configurations, effectively improving item delivery efficiency.

Another focus area is order batching and sequencing. Cals et al. (2021) apply the DRL algorithm in a warehouse combining Picker-to-Parts and Parts-to-Picker mechanisms. Well-trained DRL agent exhibited significant advantages over heuristic rules, efficiently optimizing the entire processes.

In addition, DRL has been extended to multi-agent settings for warehouse management. Christianos et al. (2020) design a multi-agent reinforcement learning (MARL) algorithm to enhance cooperation among DRL agents in a single-deep warehouse with multiple robots, achieving substantial improvements over comparative algorithms. Building on this, Krnjaic et al. (2024) develop a hierarchical MARL algorithm for order-picking problems in warehouses with both robotic and human co-workers. By optimizing the coordination of movements and actions, the proposed method significantly improved the pick rates compared to heuristic rules.

In addition to the DRL-based method, Xu et al. (2023) propose an ML-based method to address equipment planning in warehouses, demonstrating better performance than heuristic rules in single-lane depth scenarios.

Despite the advancements achieved by these studies, the potential of DRL in more complex multi-deep storage systems, where retrieval blockages and structural dependencies play crucial roles, remains largely unexplored. This gap highlights the need for solutions that integrate both local structural and global prioritization considerations, which our proposed approach aims to address.

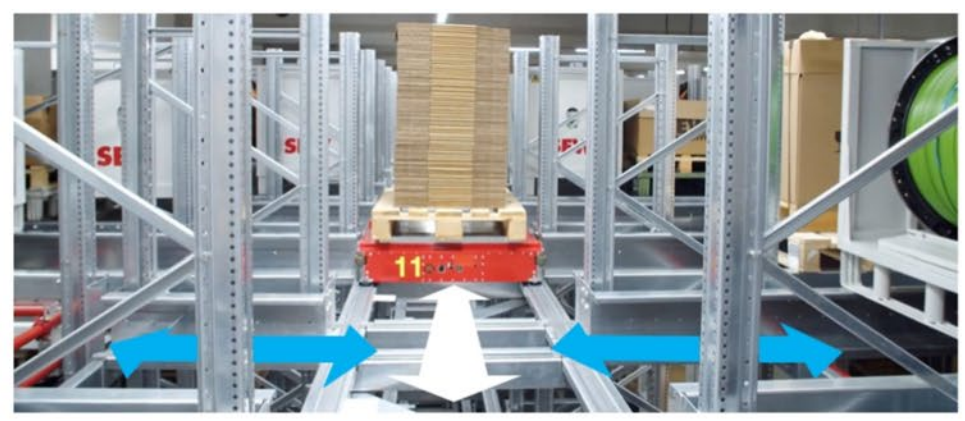

(a) A loaded shuttle in a multi-deep storage system, the cross-aisle and lanes are highlighted in white and blue, respectively (Source: Gebhardt GmbH)

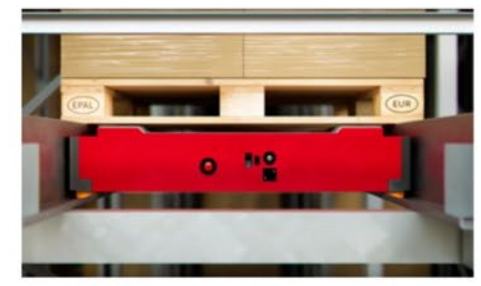

(b) Unloaded shuttle can travel under the stored items (Source: Gebhardt GmbH)

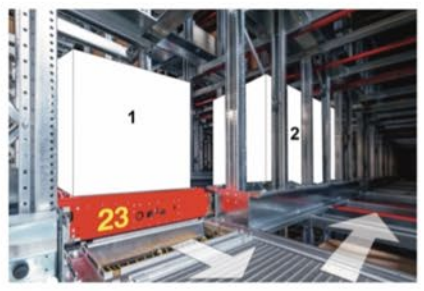

(c) Only the front item (item 1) is directly accessible (Source: Gebhardt GmbH)

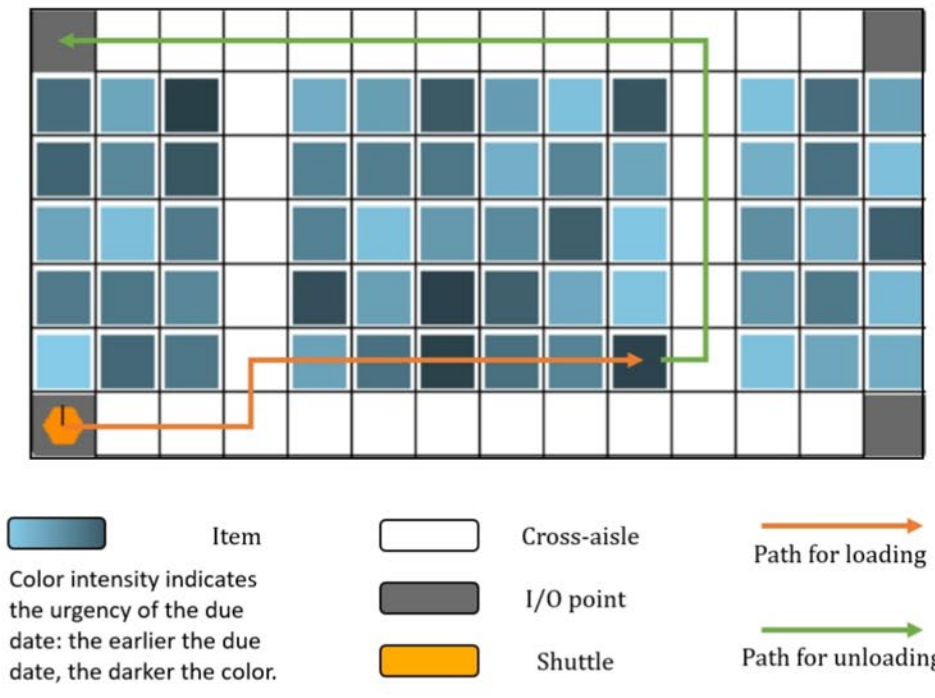

(d) Simulation of a multi-deep storage system. In the example path, the shuttle may travel beneath stored items within a lane while it is loading (i.e., when unloaded), but it may travel only in the cross-aisle or along empty lanes while unloading (i.e., when loaded).

**Fig. 4** Multi-deep storage system and the an exemplary path for retrieval process

**Table 1** Notations used in the retrieval problem

| Parameter | Description |
|---|---|
| $n$ | Total number of items |
| $i$ | Index of items |
| $x$ | Horizontal coordinate |
| $y$ | Vertical coordinate |
| $d$ | Due date |
| $t$ | Current time |

## Problem formulation

This work focuses on the retrieval process in a multi-deep AVS/RS system to minimize total tardiness. Detailed description of this warehouse is provided in Subsection 3.1.

### Problem description

This study considers a variant of AVS/RS with a shuttle, commonly known as SBS/RS. We focus on the retrieval process in a single tier of this system. The retrieval operation is performed by a single shuttle that loads items from the lanes and unloads them to the tier's I/O point. Items are heterogeneous and randomly distributed across all storage locations, and each item is associated with its individual due date. Figure 4 presents the physical details of the storage system, such as the lane arrangement and lift positions, and shows an example shuttle retrieval path in the simulation.

The notations used to describe this problem are listed in Table 1. Each item possesses an individual retrieval due date $d_i$ with $i$ represents the item index. $C_i$ refers to the $i$th item's completion time, namely the moment it reach the I/O point and is unloaded. This value is determined by the coordinates of storage location and the selected I/O point. If $C_i$ exceeds $d_i$, it incurs tardiness and is referred to as $T_i$. The objective is to minimize $T_i$ for all retrievals, i.e., total tardiness. The mathematical expression is given here:

$$\min \sum_{i=1}^{n} T_i = \min \sum_{i=1}^{n} \max(0, C_i - d_i) \tag{1}$$

To maintain a clear focus on our primary objective of minimizing total tardiness through retrieval-sequence optimization, we introduce the following assumptions:

- The warehouse is initially fully occupied by the items. We adopt this assumption for two reasons. First, our study targets retrieval-sequence optimization only, whereas a fully filled situation stresses the retrieval logic most severely. Second, the proposed RL agent satisfies the Markov property (Sutton and Barto, 2018): each decision depends solely on the current state. As retrievals proceed, partially filled layouts naturally arise as intermediate states and are handled in the same way; therefore, the full-occupancy assumption does not restrict the applicability of the method to real warehouses that often operate below the maximum capacity.
- For simplicity, the time required for all shuttle operations, including moving one step, loading, and unloading, is assumed to be 1 time unit (Christianos et al., 2020; Krnjaic et al., 2024). The focus of this work is on optimizing the retrieval sequence rather than the specific operational timings of the shuttle.
- Shuttle runs at a constant speed (Dong & Jin, 2024; Luo et al., 2021).
- Warehouse has a rectangular and symmetrical layout (Ballestín et al., 2020; D'Antonio & Chiabert, 2019).
- Once unloaded at the I/O point, item will be immediately transported out of the system (Eder, 2020).

### Operational process

As illustrated in Fig. 5, the scheduling process proceeds as follows. Initially, the shuttle enters the storage system at a randomly chosen I/O point. The DRL agent selects an item from all accessible items for the idle shuttle. The shuttle then travels to the selected item. After loading the item, the DRL agent selects an I/O point then directs the shuttle to unload it. Shuttle paths to target (selected item or I/O point) are determined by a predefined path planning algorithm, specifically Dijkstra's algorithm (Dijkstra, 1959) in the simulation. This cycle continues until all items are retrieved and unloaded.

## Proposed approach

To employ DRL, the considered retrieval process is described using a Markov decision process (MDP), $(S, A, R, \gamma, p)$. $S$ represents the observation (state) space, $A$ denotes the action space, $R$ denotes the reward function and needs to be elaborated design. $p$ and $\gamma$ are simulation-environment-related parameters and refer to the transition probability and discount factor, respectively.

DRL agent interacts with this MDP-formulated environment and receives feedback from it via the reward function. The DRL agent is represented by a neural network with novel architecture that combines GNN and Transformer. A DRL algorithm, specifically Proximal Policy Optimization (PPO) (Schulman et al., 2017), is utilized to update this neural network based on the reward from the environment.

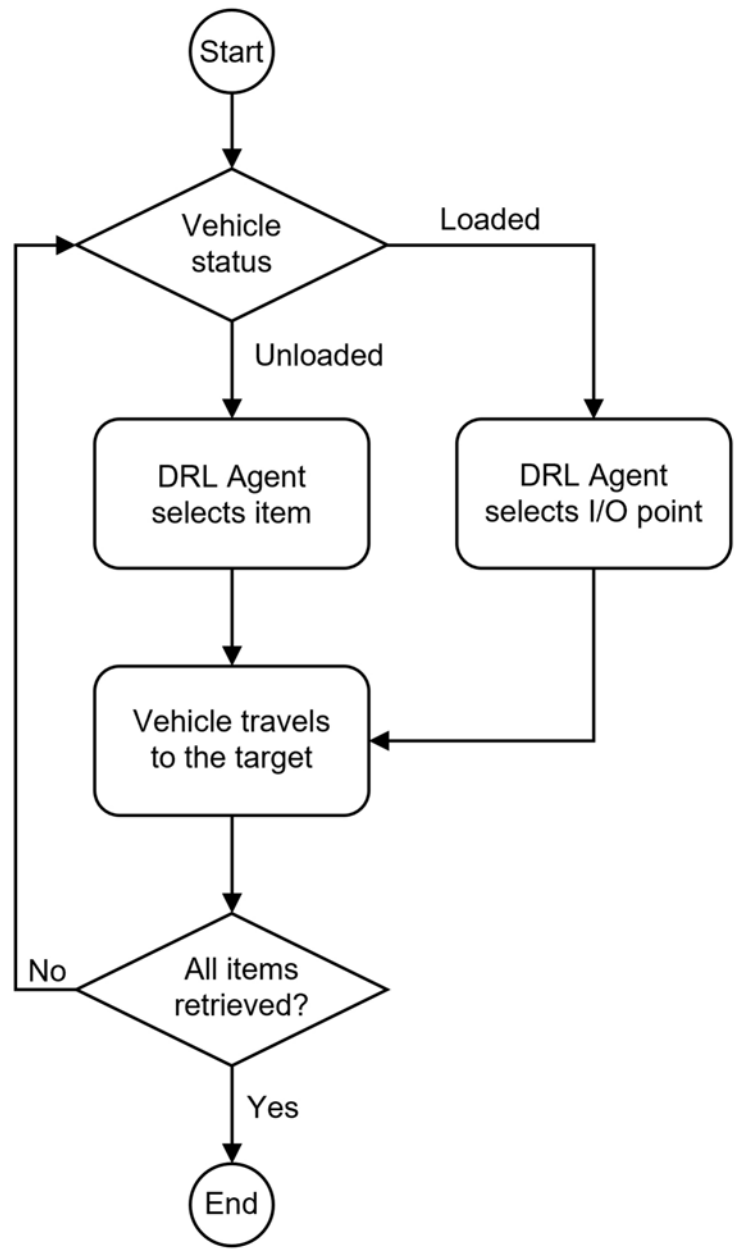

**Fig. 5** Flowchart of the scheduling process

The graph-based state representation and scalable action space are given in Sect. 4.1 and 4.2, respectively. The network architecture and the mechanism of the GNN and Transformer are presented in Sect. 4.3. Section 4.4 details the reward function design. The mechanism of PPO for networks updating is provided in Sect. 4.5.

### State representation

Building on the scheduling process, the DRL agent makes decisions about which accessible items to retrieve and which I/O points to deliver, relying on the current state of the storage system. To effectively represent not only the features of all items but also the complex structure of the multi-deep storage system, we propose a graph-based state representation. Figure 6 illustrates this representation.

In this representation, each item storage location is modeled as a node, and nodes within the same lane are connected by edges. Additionally, the cross-aisle, which allows the shuttle to access each lane and I/O points, is represented by a series of nodes and edges. The number and arrangement of these nodes correspond to the number of lanes and their respective positions, thereby capturing both the spatial layout and the connectivity within the storage system.

Each node in the proposed graph-based state is represented by an 8-dimensional feature vector. These features encode the node category, its coordinates, the due date of the stored item, and the presence of a shuttle. These node-level attributes, together with the structural information of the warehouse graph, are then processed by the DRL agent to obtain a comprehensive priority for retrieval decisions. Table 2 summarizes the meaning of each vector element. To provide a clear illustration of the proposed graph-based state representation, Fig. 7 presents the feature values of four nodes at time step 18 as an example of how each node is encoded in the state matrix. Nodes (0, 0), (1, 2), (1, 3), and (1, 4) correspond to an I/O point, an empty storage location, a highway node, and a storage location where an item is currently being carried by the shuttle, respectively.

### Action space

To support the scheduling procedure, the action space is designed for flexibility and adaptability to the warehouse state. When the shuttle is empty, the action involves selecting a directly accessible item for retrieval, which refers to items located at the front of each lane that do not require moving other items. Once the shuttle has reached and loaded the selected item, the DRL agent then chooses an appropriate I/O point for unloading. Figure 8 depicts the action space.

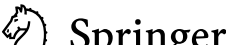

**Fig. 6** Graph-based state representation of a multi-deep storage system

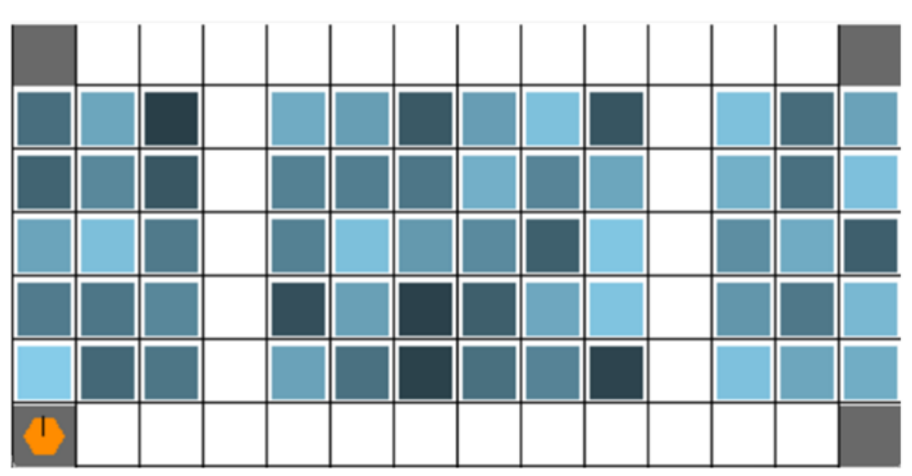

Top view of a multi-deep storage system

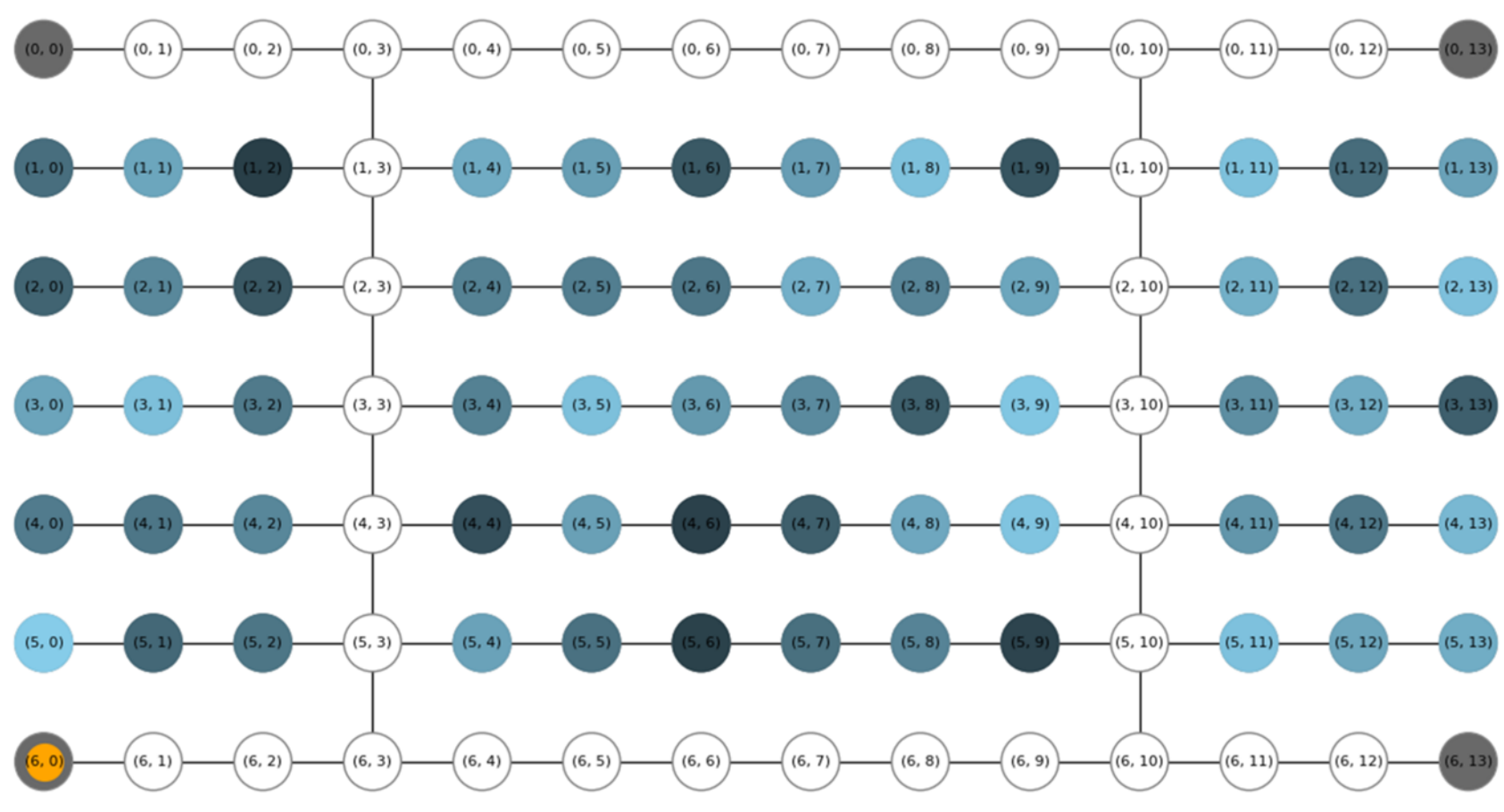

Graph-based state representation of the storage system
（Each node displays its coordinates in (y, x) format）

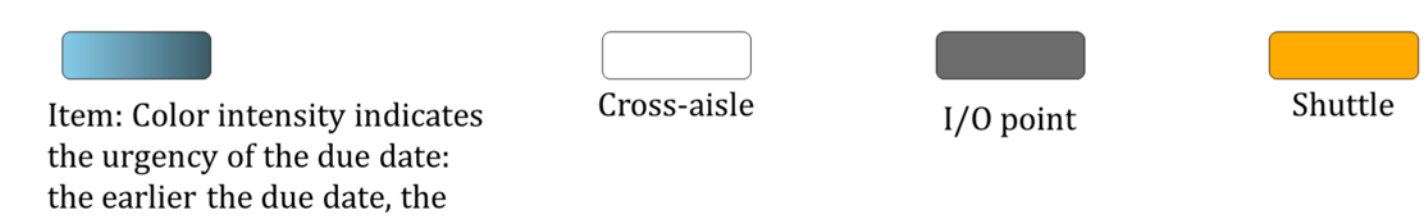

**Table 2** Node feature description

| Index | Feature value and description |
|---|---|
| 0 | Storage location indicator (1 if the node is a storage location, 0 otherwise) |
| 1 | Cross-aisle indicator (1 if the node is a cross-aisle, 0 otherwise) |
| 2 | I/O point indicator (1 if the node is an I/O point, 0 otherwise) |
| 3 | Vertical coordinate of the node $y$ |
| 4 | Horizontal coordinate of the node $x$ |
| 5 | Due date $d$ if an item is stored at this node, 0 otherwise |
| 6 | Shuttle status (1 if a shuttle carrying an item is at this node, -1 if a shuttle without an item is at this node, 0 if no shuttle is present) |
| 7 | Current time $t$ |

Since I/O points and directly accessible items differ in number, the action space is thereby flexible. Moreover, the total directly accessible items also changes as the scheduling process proceeds. When all items in a given lane are retrieved and the lane becomes empty, the count of directly accessible items decreases accordingly. This flexibility enables the proposed DRL-based approach with high scalability, allowing a well-trained agent to operate effectively across storage systems of arbitrary scale and varying configurations of lanes or I/O points.

## Neural network architecture

Choosing the optimal action from the action space, namely selecting the most appropriate item among all directly accessible items, requires the network to convert the current state to a priority score for each possible action. This task is non-trivial in multi-deep storage systems, as the priority of each directly accessible item depends not only on its relationships with other directly accessible items (global), but also on blocked items within its corresponding lane (local). To address this challenge, we elaborate a novel network architecture consisting of a GNN module and a Transformer module. The GNN encodes the features of the blocked items, while the Transformer module aggregates these local embeddings into global priorities. This division of labor

**Fig. 7** Warehouse situation at time step 18 and the feature values of node (0, 0), (1, 2), (1, 3), and (1, 4) in the corresponding state matrix

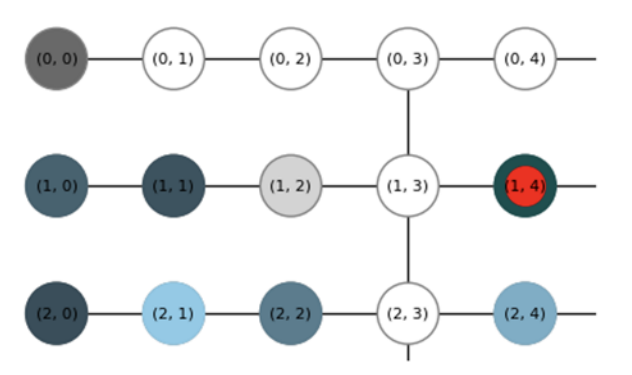

Warehouse situation at time step 18
(in graph-based representation)

| Node | Warehouse one-hot indicator | | | Coordinate | | Due date or 0 | Shuttle indicator | Current time |
|---|---|---|---|---|---|---|---|---|
| (0, 0) | 0 | 0 | 1 | 0 | 0 | 0 | 0 | 18 |
| | | | | … | | | | |
| (1, 2) | 1 | 0 | 0 | 1 | 2 | 0 | 0 | 18 |
| | | | | … | | | | |
| (1, 3) | 0 | 1 | 0 | 1 | 3 | 0 | 0 | 18 |
| | | | | … | | | | |
| (1, 4) | 1 | 0 | 0 | 1 | 4 | 810.71 | 1 | 18 |
| | | | | … | | | | |

Feature of exemplar nodes in the corresponding state matrix allows each module to play its strengths while offsetting the other's weaknesses. Using only a GNN would confine information propagation to local neighborhoods, requiring ever-deeper stacks of message-passing layers to reach across lanes. However, this inevitably leads to an over-smoothing effect, where node embeddings collapse toward a common vector and lose the distinctions needed for prioritization (Oono & Suzuki, 2019). Conversely, a single Transformer processes every blocked item as a separate token, inflating the input length and discarding the relational structure. Our ablation study in Sect. 5.2 further validates the superiority of this hybrid design. Sections 4.3.1 and 4.3.2 present the implementation details for both modules, respectively. Corresponding pseudo-codes are provided in Algorithms 3 and 4.

**Fig. 8** Action spaces for loading and unloading items

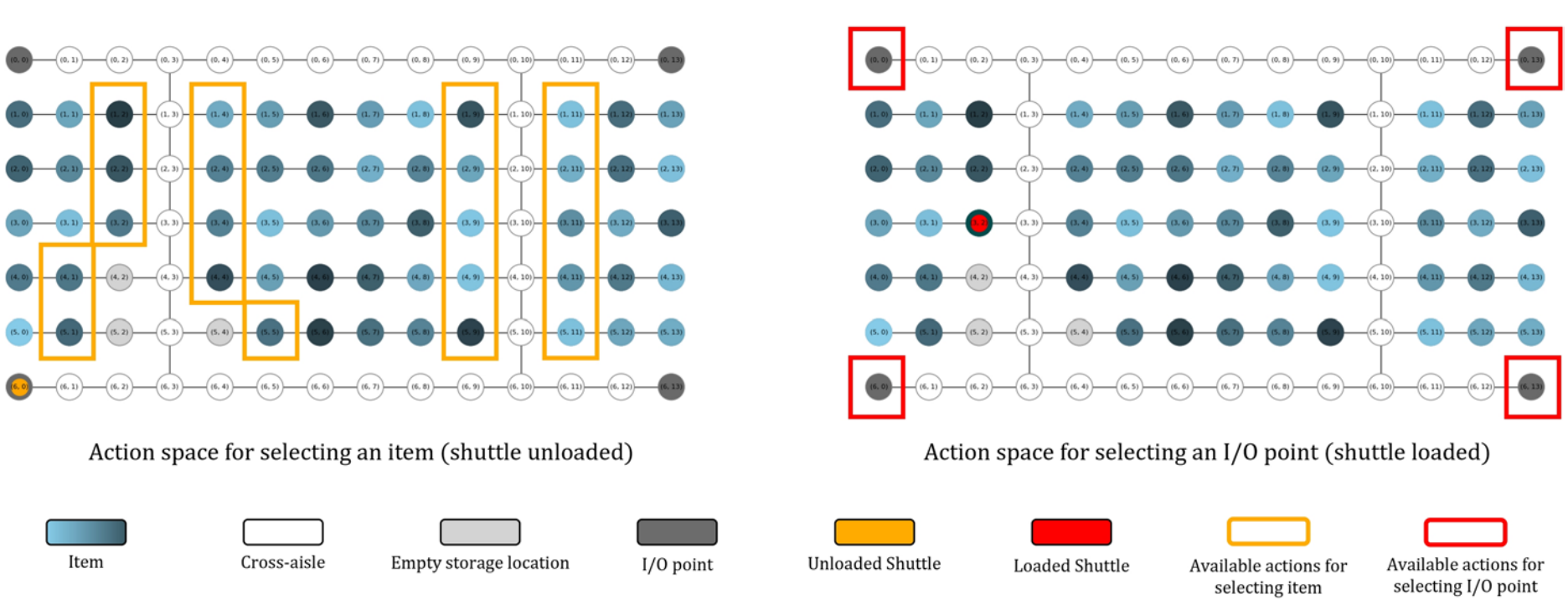

#### Mechanism of the GNN model

GNNs are designed to process graph-structured data by aggregating the features of neighboring nodes into a target node, thereby enabling the target node to effectively represent the attributes of its local neighborhood.

We adopt this principle in our proposed approach, taking each directly accessible item as a target node $i$ and embedding the features of its neighbors $\mathcal{N}(i)$, including the blocked items within the corresponding lane.

To implement this idea, we employ a specific GNN variant named GraphSAGE (Hamilton et al., 2017) containing $L$ layers, converting the raw features of each node $x_i$ into embedded features $z_i$. In each hidden layer $l$, the computation of the hidden features $h_i^l$ is mathematically expressed as follows:

$$h_i^l = \sigma(W^l \cdot AGG(\{h_j^{l-1} : j \in \mathcal{N}(i) \cup \{i\}\})) \tag{2}$$

, where $AGG(\cdot)$ refers to the aggregation function, $W^l$ denotes the parameters of the neural networks in the $l$th GNN layer. $\sigma$ is the activation function, which is selected as ReLU (Dubey et al., 2022). It is noteworthy that $h_i^0 = x_i$ and $h_i^L = z_i$. The aggregation function that we utilized is defined as:

$$AGG(\{h_j^{l-1} : j \in \mathcal{N}(i) \cup \{i\}\}) = h_i^{l-1} \parallel \frac{1}{|\mathcal{N}(i)|} \sum_{j \in \mathcal{N}(i)} h_j^{l-1} \tag{3}$$

, where $\parallel$ denotes the concatenation operation. The features of the neighbors are first averaged and then concatenated with the features of the target node, providing a richer input for the neural networks.

Figure 9 illustrates how a two-layer GraphSAGE model embeds the target node (3, 2). In the first layer, raw features of its 2-hop neighbors, including the 1-hop neighbors themselves, are aggregated to produce hidden features for each 1-hop neighbor. In the second layer, these hidden features are then combined to yield the final embedded features of the target node (3, 2).

Notably, in this process, the target node is considered part of its own 1-hop neighborhood, and consequently part of its 2-hop neighborhood (as a 1-hop neighbor of its own 1-hop neighbors). Hence, the target node's features are included in both layers' aggregation steps, ensuring that the final embedding captures information about both the target node itself and its neighbors. This comprehensive representation serves as the input to the Transformer model, enabling an efficient priority computation in the subsequent process.

**Fig. 9** GNN embedding process

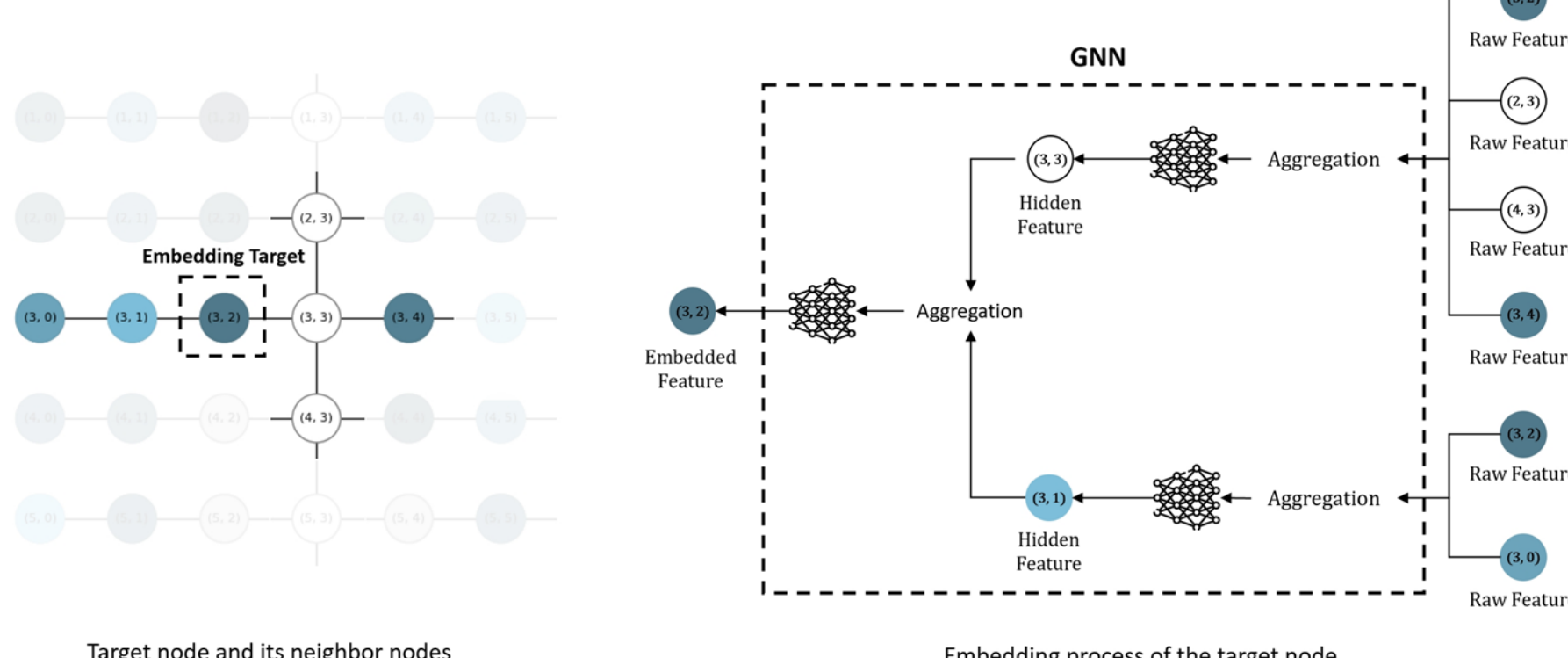

### Mechanism of the Transformer model

Since the mechanisms for selecting an item during loading and selecting an I/O point during unloading are similar, we consider the item selection process as an example to describe the working mechanism of the Transformer model.

While the GNN module encodes the local features for each directly accessible item, the subsequent priority computation process requires a global comparison among these items. To fulfill this requirement, we employ a Transformer model to convert the GNN-embedded features into priority scores. Specifically, we utilize multiple Transformer blocks to further encode these GNN-embedded features, enabling them to incorporate global dependencies from other directly accessible items. Subsequently, a Multi-Layer Perceptron (MLP) decodes these globally enriched features into the corresponding priority scores.

The core concept in the Transformer model is the Attention mechanism, which comprises three neural networks with parameters $W^Q$, $W^K$, and $W^V$. Each input vector, namely the GNN-embedded features $z_i$, is projected by these three networks into corresponding vectors:

$$
\begin{aligned}
q_i &= z_i W^Q \\
k_i &= z_i W^K \\
v_i &= z_i W^V
\end{aligned} \tag{4}
$$

The Attention operation then aggregates information across all inputs by considering their $(q, k, v)$ vectors. Formally, the output $y_i$ for the $i$th input is computed as:

$$
y_i = \sum_{j=1}^{n} \frac{\exp\left(\frac{q_i \cdot k_j}{\sqrt{d_k}}\right)}{\sum_{j'=1}^{n} \exp\left(\frac{q_i \cdot k_{j'}}{\sqrt{d_k}}\right)} v_j \tag{5}
$$

, where $d_k$ is the dimension of $k_i$ and $n$ refers to the number of inputs, namely the number of directly accessible items.

Through this process, each output embedding $y_i$ encodes all information required to compute the priority of each directly accessible item. Specifically, it integrates the local information provided by the GNN, such as the features of blocked items within the same lane, and global information captured by the Transformer, encompassing all directly accessible items across the entire warehouse. This embedding $y_i$ is then mapped to a scalar representing the priority of the node through an MLP parameterized by $W^{Prio}$:

$$
p_i = y_i W^{Prio} \tag{6}
$$

, where $p_i$ is the priority score.

During unloading, the network applies the same mechanism to calculate the priority for each I/O point. The subsequent process for selecting items or I/O points based on their priorities is detailed in Sect. 4.5.

### Reward function design

For total tardiness minimizing, a straightforward approach would be to set the reward as $-TT$. However, this leads to a sparse reward signal, hindering the agent to trace specific outcomes back to earlier decisions in the trajectory (Ng et al., 1999).

Hence, we design a dense reward function, as illustrated in Algorithm 1. For each decision point, the incremental tardiness across all items is calculated and used its negative value as the reward $r_t$ for the corresponding action $a_t$. Notably, the sum of these incremental values over all items equals the total tardiness, thus preserving the original optimization objective. Moreover, if $TT$ equals zero, indicating an optimal retrieval process, an extra positive reward $R_{opt}$ is assigned, which is defined to a constant value of 100.

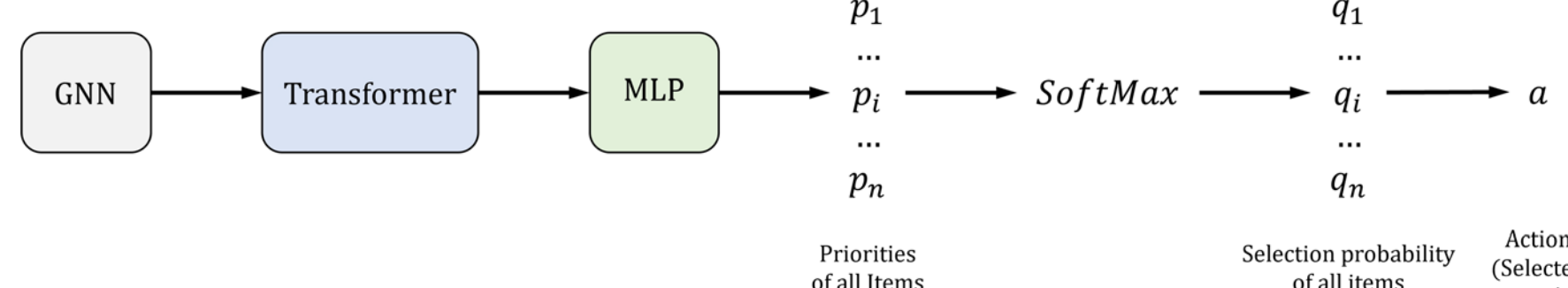

**Fig. 10** Actor network architecture

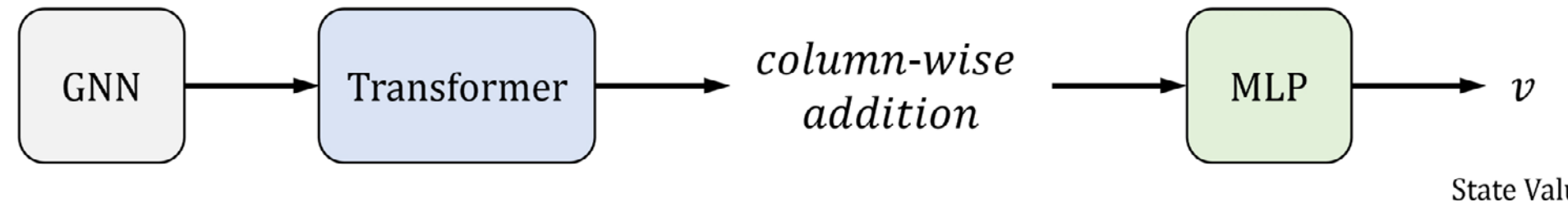

**Fig. 11** Critic network architecture

1: **Initialization:**
2: $s_t \leftarrow s_0$
3: $itemList \leftarrow \{(item_1, d_1), \ldots, (item_i, d_i), \ldots, (item_n, d_n)\}$
4: $reward \leftarrow 0$
5: $TT(t) \leftarrow 0$
6: $t \leftarrow 0$
7: **while** $itemList \neq \varnothing$ **do**
8: $a_t \leftarrow Agent(s_t)$
9: $s_{t+1} \leftarrow Env(s_t, a_t)$
10: $\text{temp} \leftarrow 0$
11: **for each** $(item, d)$ **in** $itemList$ **do**
12: $T_{item}(t) \leftarrow \max(0,\ t - d)$
13: $\text{temp} \leftarrow \text{temp} + T_{item}(t)$
14: **end for**
15: $TT(t) \leftarrow \text{temp}$
16: **if** $t = 0$ **then**
17: $reward \leftarrow reward - TT(t)$
18: **else**
19: $reward \leftarrow reward - (TT(t) - TT(t-1))$
20: **end if**
21: $s_t \leftarrow s_{t+1}$
22: **if** $a_t = \text{UNLOAD}(item_i)$ **then**
23: Remove $(item_i, d_{item_i})$ from $itemList$
24: **end if**
25: $t \leftarrow t + 1$
26: **end while**
27: **if** $TT(t) = 0$ **then**
28: $reward \leftarrow reward + R_{opt}$
29: **end if**

**Algorithm 1** Reward function

## DRL algorithm

We utilize the PPO algorithm to update the aforementioned neural networks. It adopts the Actor-Critic architecture and Algorithm 2 illustrates its working principle. The Actor outputs the probability distribution over the action space, while Critic outputs a scalar estimating the state value.

Actor and Critic are built on the architecture described in Sect. 4.3, with their specific structures depicted in Figs. 10 and 11, respectively.

In the Actor network, after the Transformer-based priority calculation, each item's priority $p_i$ is mapped to a selection probability $q_i$ using the softmax function:

$$q_i = \frac{\exp(p_i)}{\sum_{j=1}^{n} \exp(p_j)} \tag{7}$$

This ensures that the summation of all probabilities is one and items with higher priority scores obtain greater chance of selection.

In the Critic network, the priority calculation process is omitted. Instead, the Transformer output embedding is directly mapped to the state value. In particular, the embedding with dimensions of $n \times d_k$ is reduced to $1 \times d_k$ by column-wise addition, and then fed into an MLP parameterized by the $W^{value}$ to obtain the scalar state value.

**Input:** Graph $G = (\mathcal{V}, \mathcal{E})$; node features $x_i, \forall i \in \mathcal{V}$
**Output:** Priority of all directly accessible node $p_i, i \in \{1 \ldots n\}$
1: **Initialize** Actor network parameters $\theta$
2: **Initialize** Critic network parameters $\phi$
3: **Initialize** hyperparameters: learning rates $\alpha_\theta, \alpha_\phi$, discount factor $\gamma$
4: **for** each episode **do**
5: **Observe** current state $s$
6: **Calculate** distribution over action space $\mathcal{D}(\mathcal{A}) \leftarrow \text{Actor}(s; \theta)$
7: **Sample** action $a \sim \mathcal{D}(\mathcal{A})$
8: **Execute** action $a$ then **Observe** reward and next state: $(r, s') \leftarrow \text{Env}(s, a)$
9: **Calculate** state value for current state $value_s \leftarrow \text{Critic}(s; \phi)$
10: **Calculate** state value for next state: $value_{s'} \leftarrow \text{Critic}(s'; \phi)$
11: **Calculate** advantage function: $advantage \leftarrow r + \gamma \cdot value_{s'} - value_s$
12: **Update** Critic $\phi \leftarrow \phi - \alpha_\phi \cdot \nabla_\phi \text{ValueLoss}(value_s, advantage)$
13: **Update** Actor $\theta \leftarrow \theta + \alpha_\theta \cdot \nabla_\theta \text{ActorLoss}(a, s, advantage)$
14: $s \leftarrow s'$
15: **end for**

**Algorithm 2** DRL Algorithm (PPO) with Actor-Critic Architecture

## Numerical experiments

We first present the training procedure of the proposed DRL-agent. Subsequently, we compare this training procedure with two agents with only GNN and with only Transformer model, respectively. To illustrate the agent's generalization capacity, we deploy it on diverse instances absent any re-training. In addition, the aforementioned two comparative agents with different neural network architectures, together with two heuristics, are employed on these instances.

### The agent training procedure

The DRL-agent is programmed in PyTorch. The training and validation are executed in a Python-based RL environment simulating a multi-deep AVS/RS. The environment is constructed on the basis of a warehouse environment called RWARE (Papoudakis et al., 2021; Christianos et al., 2020), which is originally designed to simulate a single-deep storage system with items of identical features. All experiments

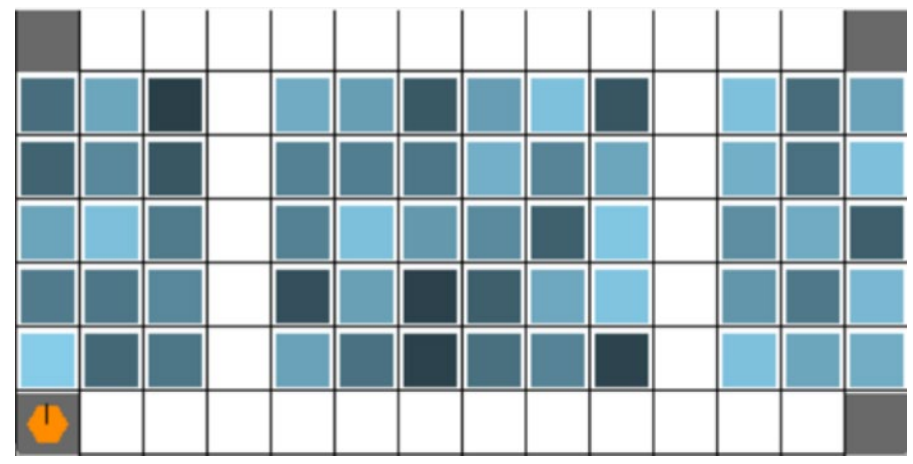

**Fig. 12** Layout of the training instance

**Table 3** Warehouse and due date parameters for training

| Warehouse dimensions | |
|---|---|
| Parameter | Value |
| Number of lanes $N_L$ | 10 |
| Lane depth $D_L$ | 3 |
| Number of cross-aisles $N_A$ | 2 |
| Number of I/O points | 4 |
| Due Date Parameters | |
| Parameter | Value |
| Due date tightness $r$ | 0.125 |
| Due date range $R$ | 0.75 |
| Lower bound | 475.5 |
| Upper bound | 1188.75 |

**Table 4** Hyperparameter settings for agent training

| Hyperparameter | Value |
|---|---|
| Number of episodes | 10000 |
| Learning rate $\eta$ | 3e-4 |
| Batch size | 1024 |
| Discount factor $\gamma$ | 0.99 |
| Cip range $\epsilon$ | 0.2 |
| Number of Transformer layers | 4 |
| Number of GNN layers | 3 |
| Optimizer | Adam |

are conducted on a PC with Intel Core i9-12900K@3.20 GHz CPU and an Nvidia RTX A4500 GPU.

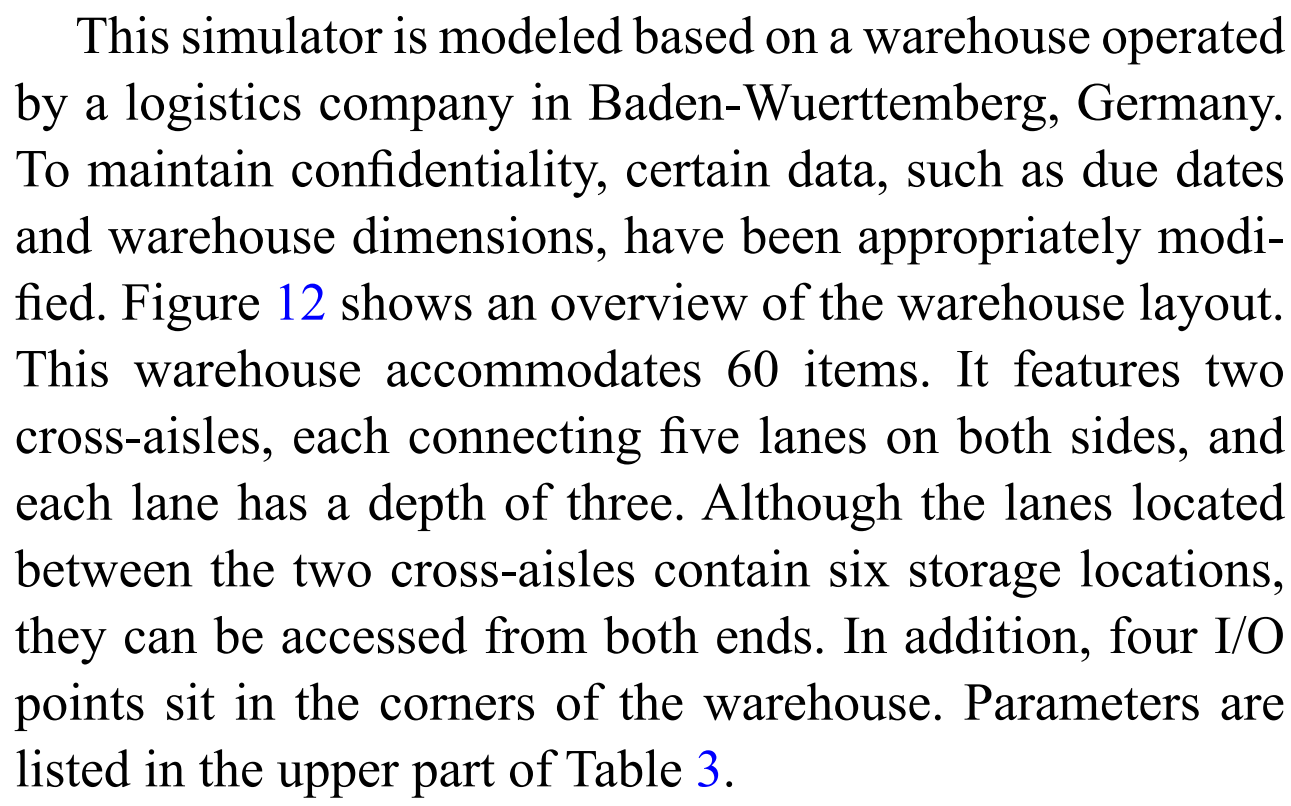

This simulator is modeled based on a warehouse operated by a logistics company in Baden-Wuerttemberg, Germany. To maintain confidentiality, certain data, such as due dates and warehouse dimensions, have been appropriately modified. Figure 12 shows an overview of the warehouse layout. This warehouse accommodates 60 items. It features two cross-aisles, each connecting five lanes on both sides, and each lane has a depth of three. Although the lanes located between the two cross-aisles contain six storage locations, they can be accessed from both ends. In addition, four I/O points sit in the corners of the warehouse. Parameters are listed in the upper part of Table 3.

In this study, each item's due date $d_i$ is generated from a widely used distribution (Potts & Van Wassenhove, 1985):

$$d_i \sim U\left(MP\left(1-r-\frac{R}{2}\right), MP\left(1-r+\frac{R}{2}\right)\right) \tag{8}$$

, where *MP* is modified total processing time, set to the makespan of the instance under the shortest travel time (STT) heuristic rule. The parameters $r$ and $R$ denote the due date tightness and the due date range, respectively. Lower part of Table 3 summarizes their specific values, and the resulting upper and lower bounds for due date distribution.

Table 4 gives the hyperparameters for training the agent. The training process is illustrated in Fig. 13, where the x-axis represents episodes, the y-axis represents the running average *TT* achieved by the agent over the last 20 episodes. Calculating the average over 20 episodes helps smooth out fluctuations in performance, making the overall training trend clearer. Initially, the agent explores the action space with limited knowledge of the environment, leading to low-quality decisions and high total tardiness. As training progresses, after approximately 1000 episodes, the agent begins to understand the structure of the warehouse and identify suitable policies, resulting in a significant decrease in total tardiness. Beyond 4000 episodes, the total tardiness decreases gradually and eventually stabilizes. The total tardiness obtained by the agent is nearly zero in the final training phase, indicating that the agent has successfully converged to an effective policy.

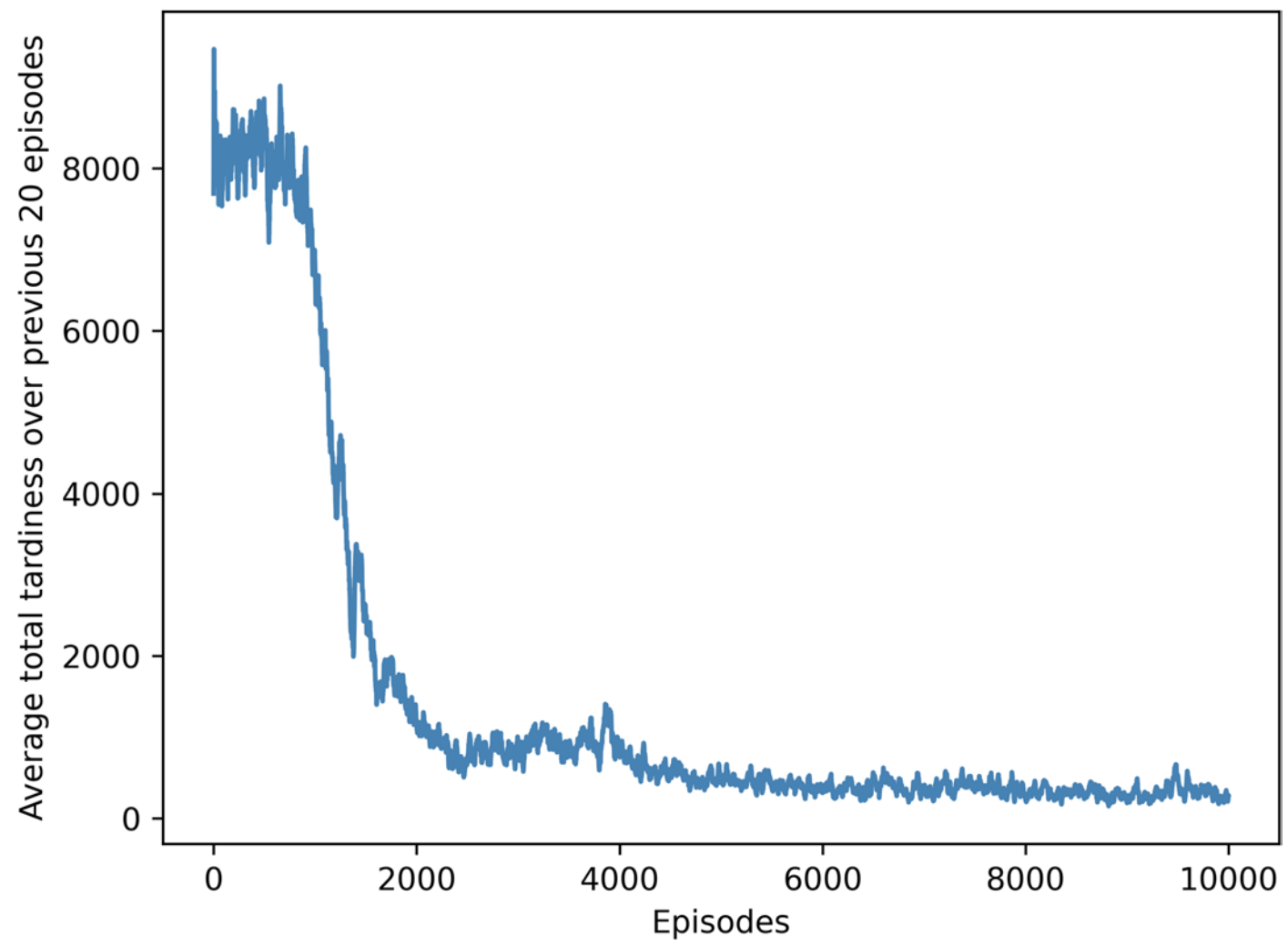

**Fig. 13** 20-episode moving average of total tardiness during training

## Ablation study

For validating the advantage of our network architecture, we have trained two comparative agents using alternative network architectures. In the first agent, the Transformer component is replaced by a GNN, while in the second agent, the GNN component is replaced by a Transformer model. This design ensures that the total number of parameters in both comparative agents is approximately the same as that of the proposed agent, eliminating the possibility that differences in performance arise owing to variations in parameter quantity rather than network structure. Furthermore, all the other training hyperparameters remain consistent with those listed in Table 4. The proposed agent and these two comparative agents are denoted as the GNN-Transformer Agent, GNN-Only Agent, and Transformer-Only Agent, respectively.

The contrast of the training processes is depicted in Fig. 14, which demonstrates the superiority of the GNN-Transformer agent. Among these three agents, the GNN-Only agent exhibits the worst performance. After a minor initial improvement in the first few episodes, the tardiness quickly stabilizes at a high level of 3000. In the later stages, the curve even shows an increasing trend due to the excessive exploration. This poor performance can be attributed to the inability of the GNN to encode global information about the warehouse, because it focuses solely on local features around each node. Consequently, it is difficult to evaluate the relative priority of items across the entire warehouse, making it difficult to establish effective retrieval strategies.

In contrast, the Transformer-Only Agent achieves a rapid reduction in the total tardiness during the first 2000 episodes. Its ability to process and compare information about all items globally enables informed early decisions. However, as the training progresses, the Transformer-Only Agent fails to further optimize its policy and fluctuates within a relatively high range of total tardiness. It is worth noting that although several approaches have successfully employed Transformer models for scheduling problems (Wang et al., 2022; Li et al., 2024; Chen et al., 2023), they are often not applied in multi-deep storage systems or even warehouse environments. This limitation arises because the Transformer model lacks a mechanism to encode warehouse structure, such as blocking relationships between items in the same lane. Consequently, it cannot effectively account for the local dependencies that are critical in multi-deep storage systems.

The proposed GNN-Transformer Agent outperforms both comparative agents by yielding a smooth training process and converging to the optimal solutions. The proposed neural network architecture combines the GNN's ability to capture local structural information with the Transformer's global prioritization capabilities, enabling an effective retrieval policy. Additionally, these three trained agents are applied to the instances in Sect. 5.3 for the generalization capability testing, and the results are given in Tables 11, 12, 13, and 14 in Appendix B. Owing to its stable convergence during training, the GNN-Transformer Agent achieved the best performance across every test instance, significantly outperforming the GNN-Only Agent and Transformer-Only Agent, further confirming the robustness of the proposed neural network architecture.

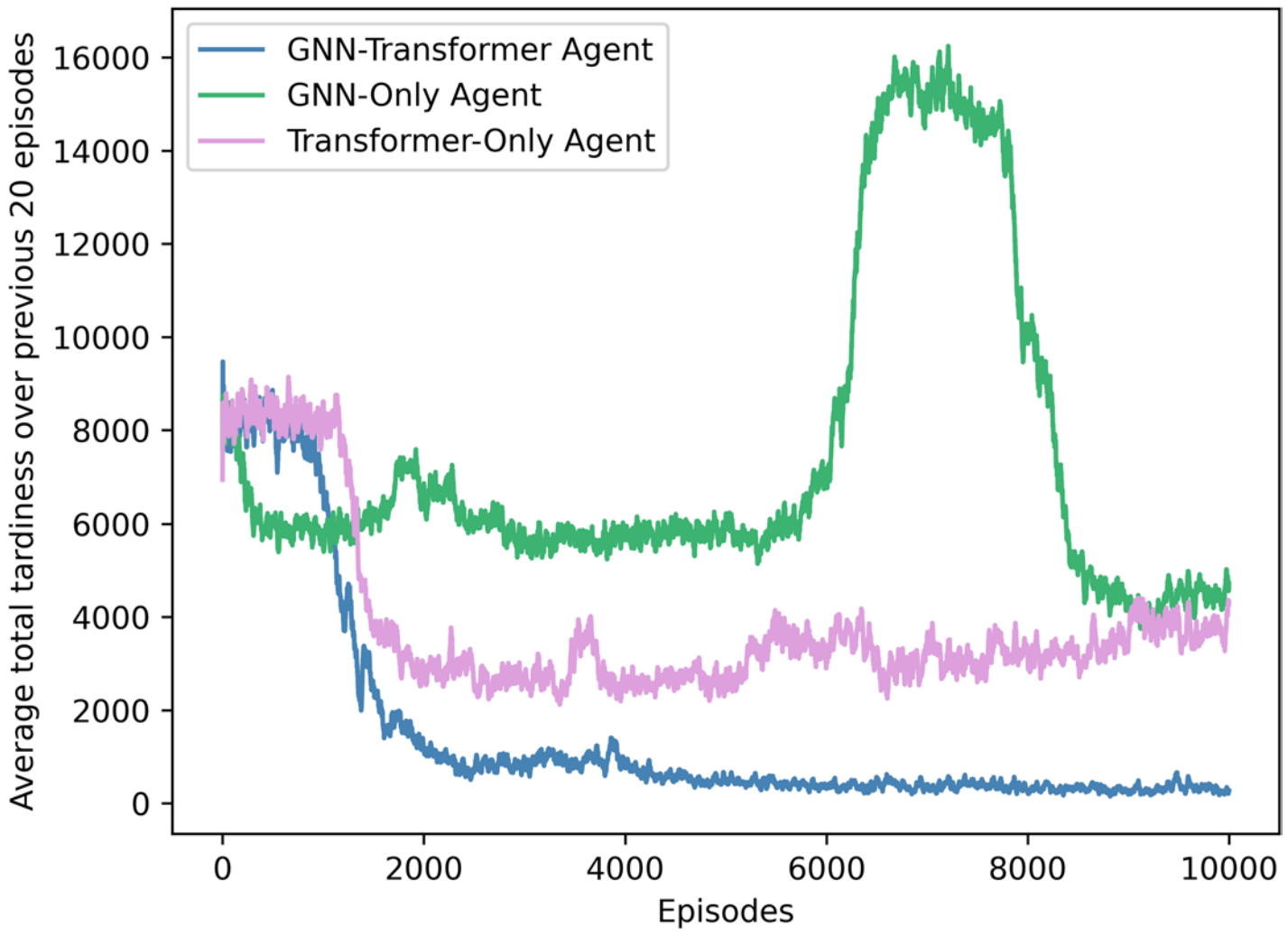

**Fig. 14** 20-episode moving average of total tardiness during training of the GNN-Transformer Agent, GNN-Only Agent, and Transformer-Only Agent

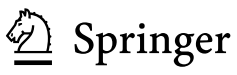

**Table 5** Warehouse dimensions for generalization capability validation

| Parameter | Value |
|---|---|
| Due date tightness $r$ | {0.125, 0.25} |
| Due date range $R$ | {0.75, 1.0} |
| Number of lanes per cross aisle $N_L$ | {6, 8, 10, 12} |
| Lane depth $D_L$ | {2, 3, 4, 5} |
| Number of cross-aisles $N_A$ | {1, 2} |

## Validation of generalization capability

To validate the generalization capability of the well-trained GNN-Transformer agent, it is then employed on 128 instances with various settings without re-training. Two heuristic rules, STT and Earliest Due Date (EDD), are also utilized to solve these various instances to evaluate the solution quality. The STT rule selects the nearest item or I/O point. The EDD rule selects the item with the earliest due date from all accessible items, and then select the nearest I/O point. Table 5 summarizes the warehouse parameters for these instances, and the corresponding layouts are given in Appendix D. According to these settings, the number of items ranges from 12 to 120.

Due to the inherent randomness in the due date generation process within the simulation environment, we independently repeat the scheduling process for all three methods 10 times for each of these 128 instances. Although the results of these 10 repetitions differ across runs, the same set of due dates is consistently used to test all three methods within each individual run, ensuring a fair comparison. It is worth noting that because of this stochastic nature of the simulation environment, methods that can adapt to changing scenarios and handle dynamic decision-making problems are required.

The mean and minimum total tardiness obtained across the 10 repetitions on the 128 instances are summarized in Tables 6, 7 8 and 9, where each table represents the set of instances with identical due date configuration. $I_i$ is the index of the validation instance. For each instance, the method with minimum total tardiness is highlighted in bold.

**Table 6** Mean and minimum total tardiness obtained by the GNN-Transformer Agent, STT, and EDD in validation instances with due date configurations of $r = 0.125$ and $R = 0.75$

| | | | | GNN-Transformer Agent | | STT | | EDD | |
|---|---|---|---|---|---|---|---|---|---|
| $D_L$ | $N_A$ | $N_L$ | $I_i$ | Mean | Min | Mean | Min | Mean | Min |
| 2 | 1 | 6 | 1 | 23.98 | 0.00 | 36.12 | 0.00 | **22.82** | 0.00 |
| | | 8 | 2 | **52.67** | 9.44 | 100.46 | 0.00 | 65.87 | 0.00 |
| | | 10 | 3 | **62.27** | 4.40 | 144.59 | 79.50 | 83.81 | 5.95 |
| | | 12 | 4 | **68.57** | 0.00 | 172.35 | 41.02 | 118.29 | 0.00 |
| | 2 | 6 | 5 | 96.31 | 0.00 | 157.74 | 24.24 | **94.87** | 0.00 |
| | | 8 | 6 | **156.34** | 0.00 | 381.44 | 155.53 | 227.95 | 120.92 |
| | | 10 | 7 | **192.81** | 0.00 | 468.27 | 213.62 | 331.10 | 91.90 |
| | | 12 | 8 | **381.38** | 0.00 | 675.05 | 344.25 | 577.69 | 322.18 |
| 3 | 1 | 6 | 9 | **29.48** | 0.00 | 101.46 | 31.07 | 82.47 | 0.00 |
| | | 8 | 10 | **54.56** | 0.00 | 278.12 | 116.48 | 209.98 | 36.53 |
| | | 10 | 11 | **35.11** | 0.00 | 274.30 | 115.39 | 277.86 | 83.30 |
| | | 12 | 12 | **51.19** | 0.00 | 532.46 | 330.42 | 498.21 | 90.47 |
| | 2 | 6 | 13 | **73.05** | 0.00 | 425.81 | 121.87 | 454.67 | 149.95 |
| | | 8 | 14 | **160.32** | 0.00 | 1003.49 | 370.71 | 1097.44 | 556.92 |
| | | 10 | 15 | **232.70** | 30.83 | 1404.18 | 1011.58 | 1615.25 | 737.80 |
| | | 12 | 16 | **300.67** | 19.15 | 2167.20 | 1357.03 | 2418.40 | 1527.32 |
| 4 | 1 | 6 | 17 | **46.85** | 0.00 | 213.68 | 50.97 | 218.40 | 0.00 |
| | | 8 | 18 | **78.97** | 0.00 | 393.25 | 141.01 | 398.95 | 86.44 |
| | | 10 | 19 | **76.30** | 0.00 | 752.90 | 426.96 | 651.37 | 212.91 |
| | | 12 | 20 | **144.94** | 0.00 | 994.21 | 231.48 | 981.05 | 298.25 |
| | 2 | 6 | 21 | **82.98** | 0.00 | 798.49 | 458.01 | 1274.28 | 470.00 |
| | | 8 | 22 | **235.61** | 0.00 | 1750.93 | 863.22 | 2797.87 | 1162.57 |
| | | 10 | 23 | **190.17** | 0.00 | 2893.25 | 2004.45 | 4246.14 | 1965.97 |
| | | 12 | 24 | **466.13** | 0.00 | 4668.36 | 3386.10 | 6398.88 | 2825.19 |
| 5 | 1 | 6 | 25 | **84.63** | 0.00 | 353.15 | 189.48 | 332.48 | 23.23 |
| | | 8 | 26 | **237.55** | 35.12 | 934.00 | 613.24 | 794.77 | 229.14 |
| | | 10 | 27 | **328.09** | 19.88 | 1379.14 | 677.95 | 1292.30 | 563.66 |
| | | 12 | 28 | **486.74** | 223.43 | 1844.85 | 1283.17 | 2053.73 | 1139.81 |
| | 2 | 6 | 29 | **309.53** | 0.00 | 1907.28 | 1218.74 | 2681.49 | 1337.36 |
| | | 8 | 30 | **611.00** | 41.11 | 3203.46 | 2041.08 | 5196.69 | 3641.98 |
| | | 10 | 31 | **849.24** | 119.12 | 5135.56 | 2456.38 | 7378.59 | 4447.20 |
| | | 12 | 32 | **1378.58** | 237.36 | 7913.43 | 5793.63 | 11828.96 | 7930.44 |

**Table 7** Mean and minimum total tardiness obtained by the GNN-Transformer Agent, STT, and EDD in validation instances with due date configurations of $r = 0.125$ and $R = 1.0$

| | | | | GNN-Transformer Agent | | STT | | EDD | |
|---|---|---|---|---|---|---|---|---|---|
| $D_L$ | $N_A$ | $N_L$ | $I_i$ | Mean | Min | Mean | Min | Mean | Min |
| 2 | 1 | 6 | 1 | **32.18** | 0.00 | 50.50 | 0.00 | 32.86 | 0.00 |
| | | 8 | 2 | **71.07** | 8.59 | 139.20 | 0.76 | 86.71 | 0.00 |
| | | 10 | 3 | **71.81** | 3.05 | 205.46 | 107.98 | 110.81 | 23.10 |
| | | 12 | 4 | **95.04** | 0.00 | 254.95 | 72.35 | 153.56 | 0.00 |
| | 2 | 6 | 5 | **121.99** | 0.00 | 260.72 | 81.51 | 127.29 | 0.00 |
| | | 8 | 6 | **197.09** | 0.00 | 535.88 | 260.07 | 235.44 | 75.96 |
| | | 10 | 7 | **278.30** | 0.00 | 706.42 | 387.92 | 399.28 | 82.25 |
| | | 12 | 8 | **505.58** | 22.25 | 989.91 | 531.08 | 649.02 | 225.00 |
| 3 | 1 | 6 | 9 | **56.45** | 0.00 | 152.32 | 54.42 | 116.24 | 0.00 |
| | | 8 | 10 | **115.57** | 0.00 | 389.69 | 165.29 | 284.63 | 56.22 |
| | | 10 | 11 | **94.92** | 0.00 | 430.14 | 212.53 | 396.30 | 100.89 |
| | | 12 | 12 | **129.83** | 0.00 | 784.24 | 474.22 | 671.17 | 116.43 |
| | 2 | 6 | 13 | **173.84** | 0.00 | 664.56 | 237.16 | 593.46 | 161.50 |
| | | 8 | 14 | **303.05** | 0.00 | 1456.96 | 707.47 | 1379.21 | 751.78 |
| | | 10 | 15 | **491.76** | 81.61 | 2164.97 | 1526.10 | 2078.76 | 1042.51 |
| | | 12 | 16 | **647.07** | 170.97 | 3237.23 | 2082.52 | 3062.98 | 2131.85 |
| 4 | 1 | 6 | 17 | **88.04** | 0.00 | 323.49 | 82.29 | 310.57 | 0.00 |
| | | 8 | 18 | **152.54** | 0.00 | 604.82 | 230.68 | 565.87 | 215.28 |
| | | 10 | 19 | **209.24** | 0.00 | 1107.78 | 701.64 | 919.82 | 392.55 |
| | | 12 | 20 | **323.45** | 0.00 | 1491.61 | 388.59 | 1380.44 | 503.00 |
| | 2 | 6 | 21 | **261.96** | 0.00 | 1249.46 | 766.79 | 1666.74 | 798.36 |
| | | 8 | 22 | **568.62** | 74.68 | 2585.01 | 1295.44 | 3497.88 | 1554.72 |
| | | 10 | 23 | **733.75** | 59.42 | 4422.63 | 3540.27 | 5412.51 | 2572.06 |
| | | 12 | 24 | **1261.91** | 6.22 | 6844.04 | 5076.20 | 8069.44 | 3965.38 |
| 5 | 1 | 6 | 25 | **154.23** | 0.00 | 536.03 | 233.66 | 474.24 | 15.64 |
| | | 8 | 26 | **374.06** | 37.16 | 1332.52 | 812.23 | 1071.88 | 363.03 |
| | | 10 | 27 | **577.48** | 11.12 | 2008.19 | 954.52 | 1799.62 | 768.56 |
| | | 12 | 28 | **903.21** | 326.26 | 2676.18 | 1758.80 | 2812.19 | 1805.32 |
| | 2 | 6 | 29 | **737.12** | 7.00 | 2889.68 | 1849.19 | 3422.35 | 1565.11 |
| | | 8 | 30 | **1504.34** | 518.82 | 4839.32 | 3588.66 | 6550.51 | 4870.86 |
| | | 10 | 31 | **1768.82** | 586.30 | 7810.25 | 3828.24 | 9328.38 | 6195.23 |
| | | 12 | 32 | **2986.30** | 1142.40 | 11745.48 | 8770.61 | 14727.72 | 10664.79 |

First, across the scenarios in Tables 6, 7, 8, and 9 the proposed GNN-Transformer agent consistently delivers strong performance under varying scales and due date configurations. In particular, across the four groups of 32 instances, this agent yields superiority over the comparative heuristics in 30, 32, 29, and 28 cases, respectively, demonstrating its robust generalization capability under diverse conditions. The advantage of the DRL-based approach is its high generalization capability, which is enabled by the novel neural network architecture. Moreover, it can simultaneously capture the local structural dependency of the warehouse and all the items from a global view. All the information is effectively computed as a comprehensive priority. By contrast, rule-based methods rely only on a limited number of features or simple predefined relationships among them.

Additionally, to verify that the superior performance of the DRL-based approach does not merely stem from simple feature computations, we further compare it against the Least Slack Time (LST) heuristic, which combines due date together with the item position into a more refined priority score and has been widely adopted for addressing scheduling problem (Baker & Trietsch, 2018). Specifically, LST selects the item with the smallest slack time $t_{slack}$:

$$t_{slack} = d - (t + \frac{l}{v_{shuttle}}) \tag{9}$$

, where $d$, $t$, $l$, and $v_{shuttle}$ refer to the due date, current time, distance from the item storage position to the I/O point, and velocity of the shuttle, respectively. As shown in Tables 15, 16, 17, and 18 in Appendix C, the LST does not change the performance superiority of the DRL agent, reinforcing that its advantage arises from its ability to learn meaningful representations beyond handcrafted rules. In addition, a Random selection baseline is included to assess the overall effectiveness of rule-based heuristics. While all rule-based methods (i.e., STT, EDD, and LST) perform significantly better than Random, the DRL agent still outperforms them by a clear margin. This confirms that its advantage is not accidental and that the baseline rules are non-trivial, further validating the robustness of the learned policy.

**Table 8** Mean and minimum total tardiness obtained by the GNN-Transformer Agent, STT, and EDD in validation instances with due date configurations of $r = 0.25$ and $R = 0.75$

| | | | | GNN-Transformer Agent | | STT | | EDD | |
|---|---|---|---|---|---|---|---|---|---|
| $D_L$ | $N_A$ | $N_L$ | $I_i$ | Mean | Min | Mean | Min | Mean | Min |
| 2 | 1 | 6 | 1 | 53.76 | 12.40 | 73.45 | 10.61 | **43.06** | 0.00 |
| | | 8 | 2 | 133.73 | 15.44 | 191.47 | 44.34 | **112.74** | 0.99 |
| | | 10 | 3 | **158.65** | 39.41 | 273.21 | 129.71 | 160.83 | 38.45 |
| | | 12 | 4 | **226.38** | 15.96 | 340.53 | 186.55 | 229.62 | 27.93 |
| | 2 | 6 | 5 | 235.53 | 15.49 | 353.96 | 168.85 | **215.33** | 32.19 |
| | | 8 | 6 | **439.70** | 184.18 | 749.71 | 492.61 | 548.57 | 382.57 |
| | | 10 | 7 | **547.96** | 67.20 | 948.97 | 660.81 | 750.80 | 405.53 |
| | | 12 | 8 | **980.68** | 339.90 | 1367.19 | 839.62 | 1312.01 | 793.13 |
| 3 | 1 | 6 | 9 | **97.65** | 0.00 | 218.06 | 121.80 | 163.89 | 19.86 |
| | | 8 | 10 | **214.44** | 10.53 | 508.31 | 240.85 | 368.08 | 83.96 |
| | | 10 | 11 | **159.98** | 0.00 | 614.13 | 330.71 | 536.88 | 266.09 |
| | | 12 | 12 | **331.61** | 66.85 | 1047.24 | 648.74 | 872.88 | 399.98 |
| | 2 | 6 | 13 | **349.57** | 77.74 | 908.83 | 395.92 | 856.51 | 453.31 |
| | | 8 | 14 | **710.46** | 420.38 | 1938.07 | 1093.45 | 2012.15 | 1253.12 |
| | | 10 | 15 | **993.27** | 268.85 | 2852.73 | 2462.37 | 2897.55 | 2061.91 |
| | | 12 | 16 | **1603.11** | 943.51 | 4340.82 | 3299.66 | 4613.47 | 3672.67 |
| 4 | 1 | 6 | 17 | **165.77** | 0.00 | 419.68 | 109.56 | 403.05 | 0.00 |
| | | 8 | 18 | **287.36** | 25.70 | 806.58 | 378.58 | 726.26 | 286.82 |
| | | 10 | 19 | **354.45** | 18.30 | 1432.50 | 988.18 | 1170.01 | 528.10 |
| | | 12 | 20 | **643.93** | 24.00 | 2031.57 | 1127.47 | 1799.79 | 905.08 |
| | 2 | 6 | 21 | **513.82** | 0.00 | 1761.38 | 1137.87 | 2243.75 | 1250.01 |
| | | 8 | 22 | **1402.09** | 425.19 | 3456.87 | 2107.66 | 4729.55 | 2836.74 |
| | | 10 | 23 | **1658.24** | 367.79 | 5736.38 | 4400.95 | 7194.92 | 3483.25 |
| | | 12 | 24 | **2733.56** | 671.81 | 9057.90 | 7061.07 | 10986.70 | 6358.61 |
| 5 | 1 | 6 | 25 | **290.92** | 23.66 | 767.71 | 454.50 | 645.07 | 204.21 |
| | | 8 | 26 | **682.10** | 216.41 | 1751.33 | 1253.31 | 1418.15 | 768.61 |
| | | 10 | 27 | **995.90** | 402.55 | 2631.66 | 1567.43 | 2299.64 | 1548.91 |
| | | 12 | 28 | **1718.27** | 1054.19 | 3679.90 | 2908.54 | 3644.36 | 2558.12 |
| | 2 | 6 | 29 | **1376.10** | 585.96 | 3849.12 | 2872.13 | 4547.84 | 2834.45 |
| | | 8 | 30 | **2776.26** | 1626.59 | 6515.93 | 5329.94 | 8871.46 | 6975.28 |
| | | 10 | 31 | **3542.06** | 1892.70 | 10264.59 | 6414.12 | 12353.86 | 8683.07 |
| | | 12 | 32 | **6325.23** | 4086.11 | 15401.23 | 12061.08 | 20117.73 | 15128.73 |

Furthermore, in these four sets, the proposed GNN-Transformer Agent successfully finds the optimal solution, that is, a schedule with zero total tardiness, in 22, 17, 4, and 3 instances, respectively. The higher occurrence of optimal solutions in Tables 6 and 7 can be attributed to the fact that these instances are generated with a smaller $r$ parameter, resulting in a lower overall tightness. In contrast, the instances in Tables 8 and 9 exhibit higher tightness and urgency levels owing to their larger $r$ values. Moreover, given the NP-hard nature of the objective of minimizing total tardiness (Jiang et al., 2021; Cebi et al., 2020; Graham et al., 1979; Lenstra & Kan, 1979) and scheduling problems in multi-deep storage systems (Yang et al., 2023), determining whether an optimal solution with zero total tardiness exists for each instance is infeasible in practice.

To further investigate the factors influencing the generalization capability of the well-trained GNN-Transformer agent, we calculate its improvement ratio of the mean total tardiness relative to the best solutions obtained using the heuristic rules. The improvement ratio is computed as:

$$Improvement = \frac{\min(TT_{EDD}, TT_{STT}) - TT_{Agent}}{\min(TT_{EDD}, TT_{STT})} \times 100\% \qquad (10)$$

, where $TT_{Agent}$, $TT_{EDD}$, and $TT_{STT}$ represent the total tardiness obtained by the GNN-Transformer Agent, the EDD rule and STT rule, respectively.

**Table 9** Mean and minimum total tardiness obtained by the GNN-Transformer Agent, STT, and EDD in validation instances with due date configurations of $r = 0.25$ and $R = 1.0$

| | | | | GNN-Transformer Agent | | STT | | EDD | |
|---|---|---|---|---|---|---|---|---|---|
| $D_L$ | $N_A$ | $N_L$ | $I_i$ | Mean | Min | Mean | Min | Mean | Min |
| 2 | 1 | 6 | 1 | 61.35 | 11.98 | 92.93 | 3.82 | **55.31** | 5.71 |
| | | 8 | 2 | 140.79 | 28.59 | 240.56 | 49.79 | **135.04** | 13.66 |
| | | 10 | 3 | **173.40** | 35.55 | 342.59 | 147.40 | 189.04 | 55.60 |
| | | 12 | 4 | **250.60** | 7.28 | 443.70 | 199.48 | 262.33 | 46.22 |
| | 2 | 6 | 5 | 259.84 | 0.00 | 464.69 | 223.46 | **211.94** | 46.10 |
| | | 8 | 6 | 479.79 | 142.92 | 935.84 | 612.05 | **447.18** | 227.95 |
| | | 10 | 7 | **602.21** | 72.37 | 1244.97 | 858.65 | 688.22 | 310.99 |
| | | 12 | 8 | **1091.18** | 237.55 | 1737.16 | 985.00 | 1119.57 | 668.57 |
| 3 | 1 | 6 | 9 | **113.46** | 1.02 | 281.74 | 155.40 | 208.75 | 21.62 |
| | | 8 | 10 | **276.38** | 54.05 | 645.74 | 325.06 | 454.73 | 125.05 |
| | | 10 | 11 | **233.13** | 24.84 | 799.87 | 402.97 | 671.61 | 302.26 |
| | | 12 | 12 | **485.78** | 122.77 | 1345.75 | 823.75 | 1073.48 | 485.84 |
| | 2 | 6 | 13 | **488.61** | 0.00 | 1217.04 | 637.97 | 1000.86 | 540.81 |
| | | 8 | 14 | **886.83** | 142.68 | 2493.01 | 1573.15 | 2126.47 | 1355.81 |
| | | 10 | 15 | **1226.75** | 267.46 | 3813.84 | 3263.58 | 3297.04 | 2179.42 |
| | | 12 | 16 | **1938.80** | 1081.17 | 5655.42 | 4201.74 | 4983.42 | 3948.07 |
| 4 | 1 | 6 | 17 | **231.30** | 0.00 | 559.96 | 137.29 | 519.60 | 0.00 |
| | | 8 | 18 | **416.87** | 52.60 | 1050.67 | 549.29 | 924.14 | 425.76 |
| | | 10 | 19 | **555.44** | 26.34 | 1880.38 | 1392.62 | 1509.43 | 786.74 |
| | | 12 | 20 | **929.32** | 56.63 | 2708.00 | 1629.11 | 2269.22 | 1243.03 |
| | 2 | 6 | 21 | **858.27** | 64.48 | 2352.17 | 1534.49 | 2677.33 | 1610.12 |
| | | 8 | 22 | **1768.82** | 454.02 | 4547.89 | 2884.63 | 5349.06 | 3296.54 |
| | | 10 | 23 | **2570.29** | 932.45 | 7685.24 | 6497.57 | 8337.83 | 4271.35 |
| | | 12 | 24 | **4211.94** | 1103.05 | 11876.39 | 9194.53 | 12440.55 | 7377.14 |
| 5 | 1 | 6 | 25 | **447.08** | 25.64 | 1014.40 | 562.90 | 812.24 | 192.95 |
| | | 8 | 26 | **893.43** | 454.56 | 2218.92 | 1469.72 | 1745.79 | 893.22 |
| | | 10 | 27 | **1393.92** | 566.85 | 3503.57 | 2075.28 | 2869.00 | 1820.77 |
| | | 12 | 28 | **2265.49** | 1117.73 | 4887.47 | 3529.33 | 4523.17 | 3110.20 |
| | 2 | 6 | 29 | **2111.50** | 935.21 | 5046.06 | 3885.46 | 5292.53 | 3169.43 |
| | | 8 | 30 | **3945.59** | 2361.35 | 8645.77 | 6857.97 | 10072.31 | 7716.91 |
| | | 10 | 31 | **5086.61** | 1943.47 | 13666.75 | 8481.11 | 14634.19 | 11707.98 |
| | | 12 | 32 | **8507.96** | 5479.60 | 20457.41 | 15884.89 | 22897.87 | 18635.64 |

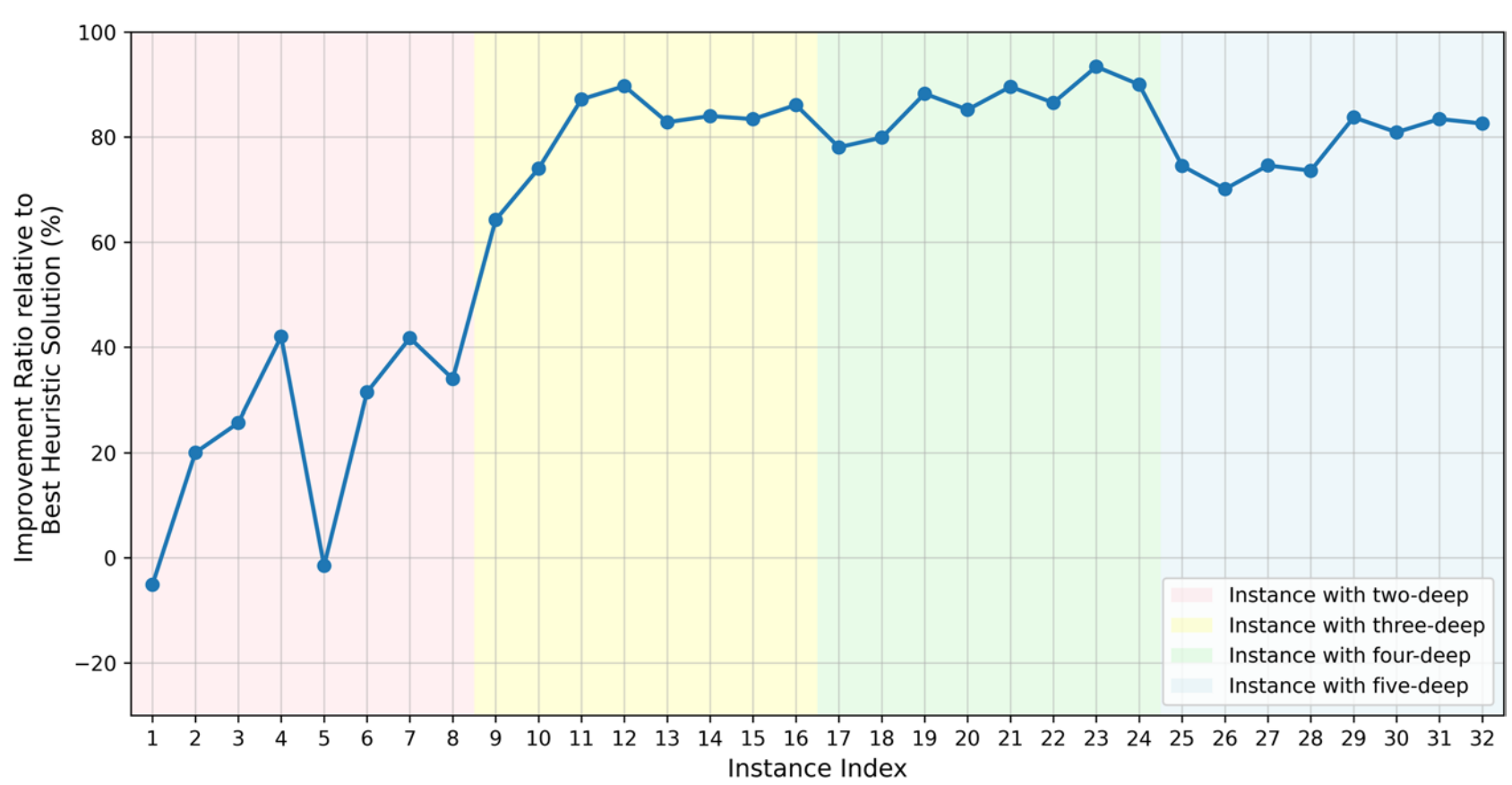

**Fig. 15** Improvement ratio relative to best heuristic solution over validation instances with due date configurations of $r = 0.125$ and $R = 0.75$

The results are depicted in Figs. 15, 16, 17 and 18, where each corresponds to one set of validation instances. In each Figure, the instances with two, three, four, and five depths are marked with red, yellow, green, and blue, respectively.

For two-deep storage, the performance improvement achieved by the DRL approach is relatively marginal, and in some instances, even slightly inferior to the heuristic rules. This can be attributed to the simple structure and relatively few items in the two-deep configurations, which limits the potential optimization space for the DRL model. However, even in cases in which the DRL approach does not outperform heuristics, the results remain competitive.

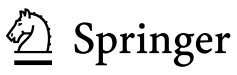

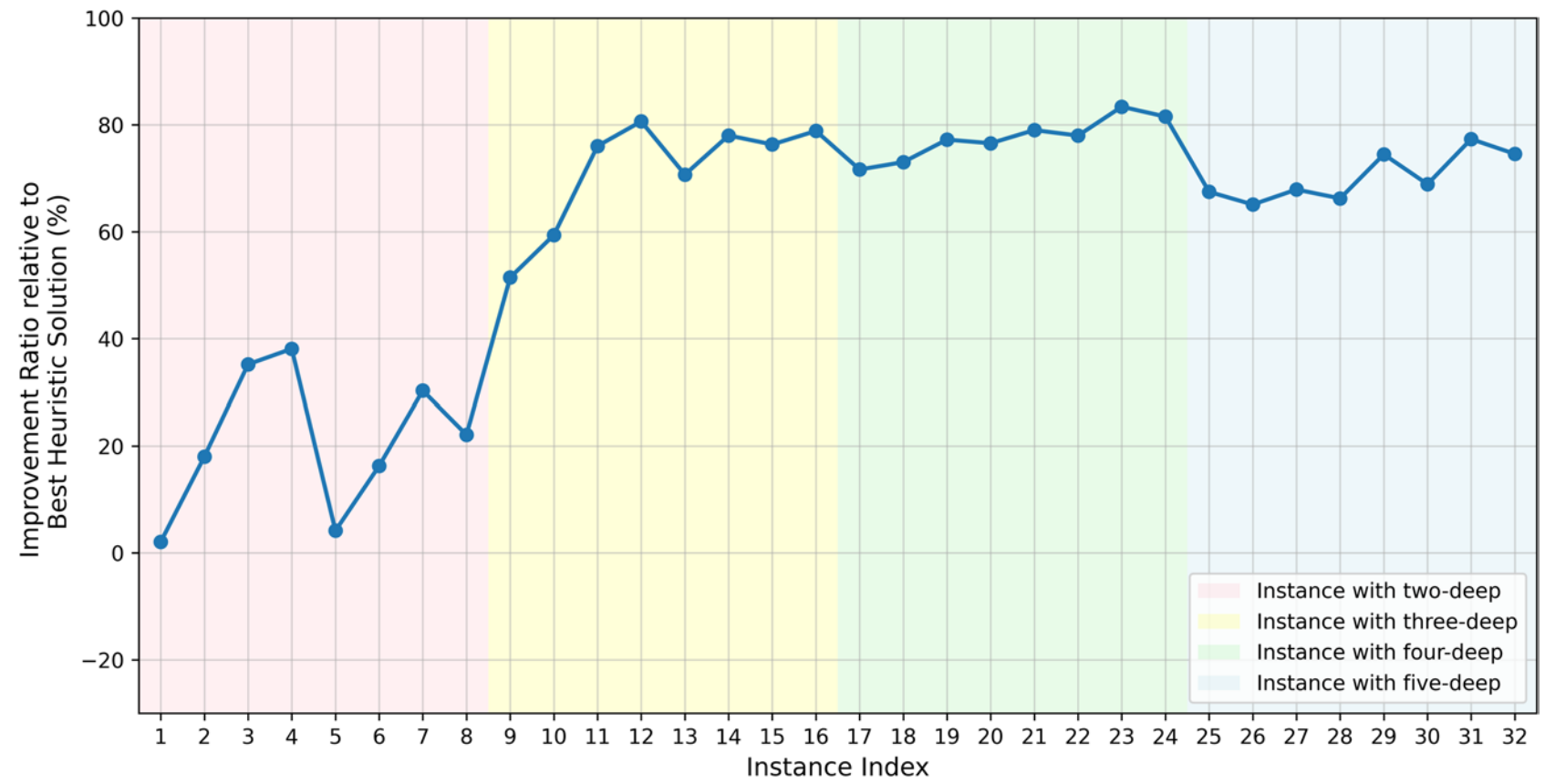

**Fig. 16** Improvement ratio relative to best heuristic solution over validation instances with due date configurations of $r = 0.125$ and $R = 1.0$

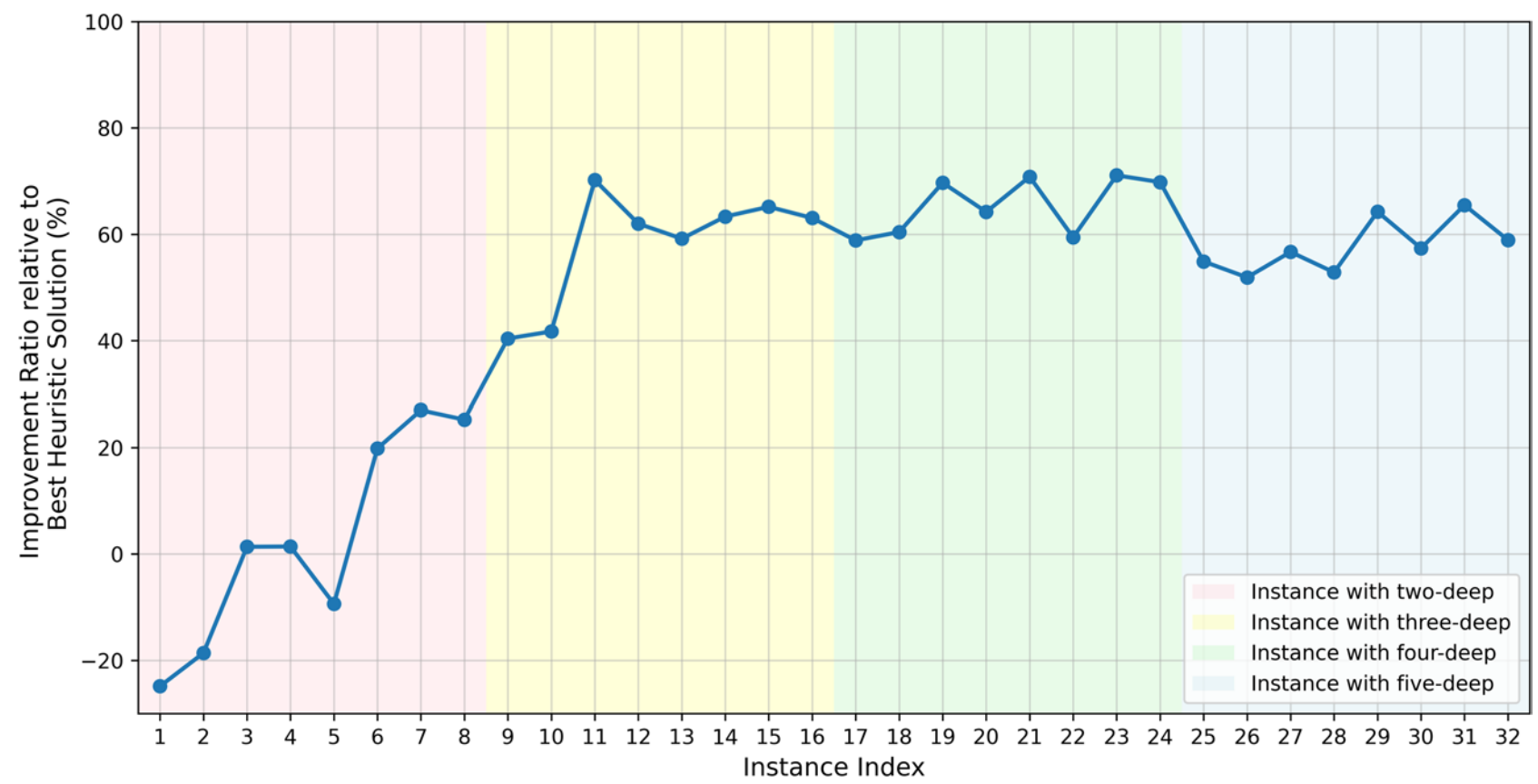

**Fig. 17** Improvement ratio relative to best heuristic solution over validation instances with due date configurations of $r = 0.25$ and $R = 0.75$

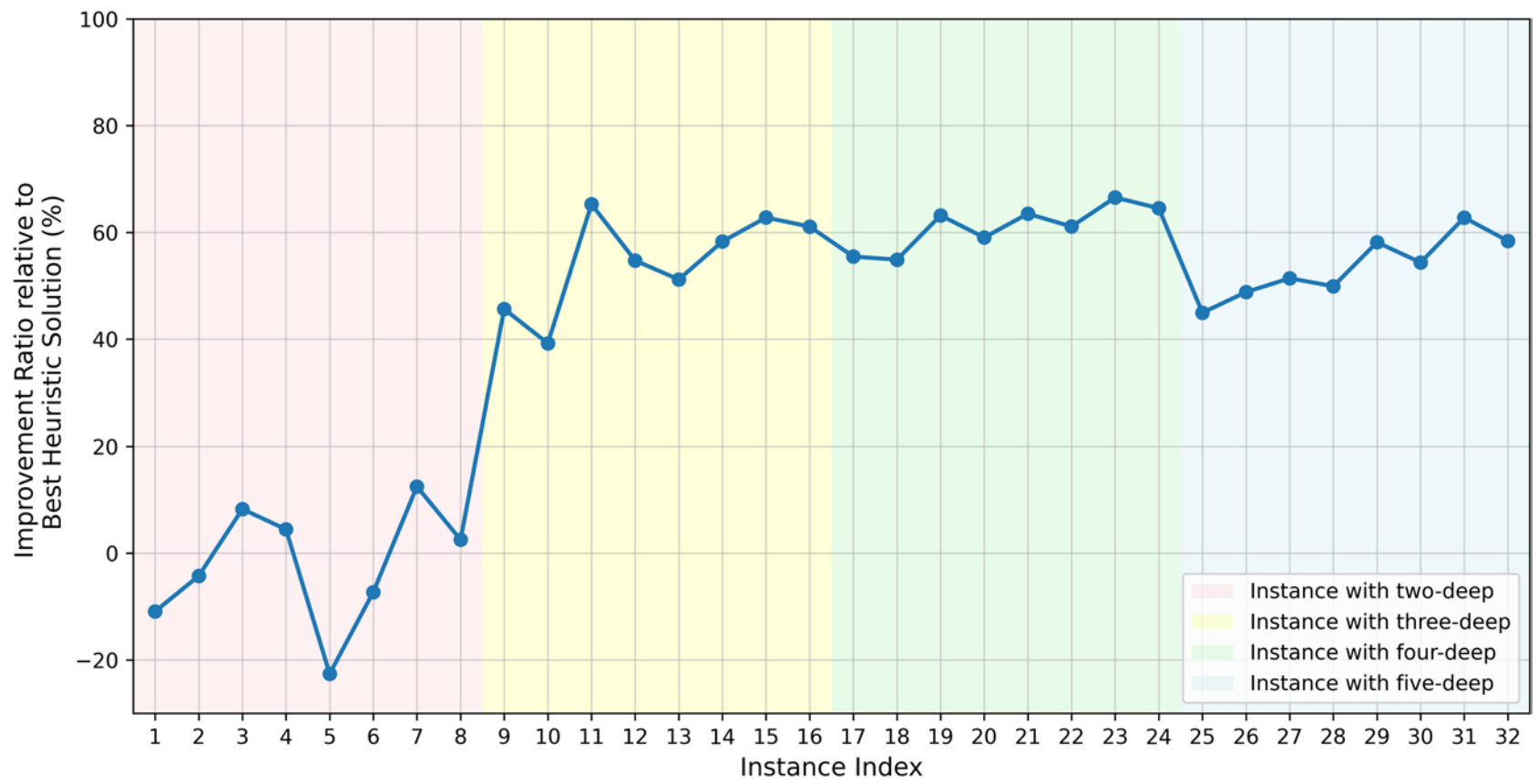

**Fig. 18** Improvement ratio relative to best heuristic solution over validation instances with due date configurations of $r = 0.25$ and $R = 1.0$

As the storage depth increases, the structural complexity introduced by multi-deep storage systems becomes more pronounced, and the number of items also increases, leading to a larger decision space. In this setting, the well-trained DRL agent demonstrates a significant advantage over the heuristic rules. Notably, in four- and five-deep configurations, the DRL approach consistently achieves improvements in the range of 60% to 80% over heuristic rules, highlighting its effectiveness in handling complex retrieval scenarios.

Moreover, the proposed DRL approach yields superior performance on instances with lower due date tightness

$r$. Across both sets with $r = 0.125$, the agent achieves improvements exceeding 60% on 24 and 22 instances, respectively. This can be attributed to the increased flexibility in scheduling decisions caused by the reduced tightness. In such scenarios, the agent has more opportunities to balance retrieval priorities and minimize unnecessary tardiness without being constrained by multiple urgent due dates. This trend suggests that when due dates are more relaxed, the DRL agent can better leverage its decision-making capabilities to optimize retrieval sequences and thus achieve lower total tardiness.

Table 10 lists the computational time each approach takes at different scales, where $N_{item}$ refers to the number of items in the corresponding instance. The reported value is the average runtime over all due date configurations, since instance scale exerts the main influence on runtime. The well-trained agent solves the largest instance in 3579.38 ms, i.e. within four seconds. Although rule-based methods require less computational time because of their simplicity, their solution quality is significantly inferior to that generated by the trained DRL agent, as it demonstrated above. Moreover, for each individual retrieval decision, the proposed agent requires only a few additional milliseconds to deliver substantially better performance, which reveals its efficiency.

It is equally notable that in the largest instance, the state graph comprises 176 nodes, each with an 8-dimensional feature vector, effectively forming a $176 \times 8$ node feature matrix. Of these nodes, 120 correspond to stored items, which in turn require $120 \times 2$ sequential decisions (one for loading and one for unloading). Meanwhile the solution space is extremely large; even if we consider only the loading steps, the total number of possible retrieval sequences is

$$\frac{N_{item}!}{((N_A \times N_L)!)^{D_L}} = \frac{120!}{((2 \times 12))!)^5} \approx 7.28 \times 10^{79} \tag{11}$$

**Table 10** Average computational time per instance and per retrieval (in millisecond) of the GNN-Transformer Agent, STT, and EDD on the validation instances across all due-date configurations

| | | | | | GNN-Transformer Agent | | STT | | EDD | |
|---|---|---|---|---|---|---|---|---|---|---|
| $D_L$ | $N_A$ | $N_L$ | $I_i$ | $N_{item}$ | per instance | per retrieval | per instance | per retrieval | per instance | per retrieval |
| 2 | 1 | 6 | 1 | 12 | 59.90 | 4.99 | 27.62 | 2.30 | 26.38 | 2.20 |
| | | 8 | 2 | 16 | 90.93 | 5.68 | 39.79 | 2.49 | 40.66 | 2.54 |
| | | 10 | 3 | 20 | 125.32 | 6.27 | 57.02 | 2.85 | 59.38 | 2.97 |
| | | 12 | 4 | 24 | 164.11 | 6.84 | 76.50 | 3.19 | 81.13 | 3.38 |
| | 2 | 6 | 5 | 24 | 190.04 | 7.92 | 93.91 | 3.91 | 95.86 | 3.99 |
| | | 8 | 6 | 32 | 292.05 | 9.13 | 150.45 | 4.70 | 153.03 | 4.78 |
| | | 10 | 7 | 40 | 414.34 | 10.36 | 220.47 | 5.51 | 241.21 | 6.03 |
| | | 12 | 8 | 48 | 573.32 | 11.94 | 324.87 | 6.77 | 330.41 | 6.88 |
| 3 | 1 | 6 | 9 | 18 | 114.70 | 6.37 | 57.07 | 3.17 | 55.87 | 3.10 |
| | | 8 | 10 | 24 | 174.72 | 7.28 | 86.37 | 3.60 | 89.43 | 3.73 |
| | | 10 | 11 | 30 | 243.93 | 8.13 | 125.33 | 4.18 | 130.49 | 4.35 |
| | | 12 | 12 | 36 | 328.78 | 9.13 | 183.45 | 5.10 | 182.37 | 5.07 |
| | 2 | 6 | 13 | 36 | 386.77 | 10.74 | 221.07 | 6.14 | 225.87 | 6.27 |
| | | 8 | 14 | 48 | 608.46 | 12.68 | 368.42 | 7.68 | 378.51 | 7.89 |
| | | 10 | 15 | 60 | 861.03 | 14.35 | 561.96 | 9.37 | 564.88 | 9.41 |
| | | 12 | 16 | 72 | 1188.82 | 16.51 | 800.80 | 11.12 | 798.37 | 11.09 |
| 4 | 1 | 6 | 17 | 24 | 192.34 | 8.01 | 104.23 | 4.34 | 104.43 | 4.35 |
| | | 8 | 18 | 32 | 292.94 | 9.15 | 158.47 | 4.95 | 177.53 | 5.55 |
| | | 10 | 19 | 40 | 419.26 | 10.48 | 233.51 | 5.84 | 251.55 | 6.29 |
| | | 12 | 20 | 48 | 568.66 | 11.85 | 339.99 | 7.08 | 355.03 | 7.40 |
| | 2 | 6 | 21 | 48 | 666.71 | 13.89 | 446.39 | 9.30 | 444.82 | 9.27 |
| | | 8 | 22 | 64 | 1063.22 | 16.61 | 724.59 | 11.32 | 730.27 | 11.41 |
| | | 10 | 23 | 80 | 1564.10 | 19.55 | 1104.29 | 13.80 | 1134.84 | 14.19 |
| | | 12 | 24 | 96 | 2138.43 | 22.28 | 1585.32 | 16.51 | 1593.68 | 16.60 |
| 5 | 1 | 6 | 25 | 30 | 289.36 | 9.65 | 172.88 | 5.76 | 176.83 | 5.89 |
| | | 8 | 26 | 40 | 453.26 | 11.33 | 280.35 | 7.01 | 281.74 | 7.04 |
| | | 10 | 27 | 50 | 644.66 | 12.89 | 413.61 | 8.27 | 426.61 | 8.53 |
| | | 12 | 28 | 60 | 877.47 | 14.62 | 567.14 | 9.45 | 607.43 | 10.12 |
| | 2 | 6 | 29 | 60 | 1078.30 | 17.97 | 774.50 | 12.91 | 795.47 | 13.26 |
| | | 8 | 30 | 80 | 1752.04 | 21.90 | 1297.96 | 16.22 | 1305.78 | 16.32 |
| | | 10 | 31 | 100 | 2532.81 | 25.33 | 1959.04 | 19.59 | 2014.15 | 20.14 |
| | | 12 | 32 | 120 | 3579.38 | 29.83 | 2862.31 | 23.85 | 2824.77 | 23.54 |

because only the front-most item in each of the 24 lanes (with a depth of 5) is selectable at any time. These numbers underscore the DRL agent's ability to process high-dimensional inputs and navigate an astronomically large solution space, delivering scalable, real-time retrieval strategies where traditional methods are infeasible.

## Discussion

Following extensive numerical experiments across multiple scenarios, the proposed DRL-based approach exhibits notable scalability and consistent efficacy. Discussion on its application is presented in this section.

### Application potential

The proposed DRL-based approach exhibits promising potential for real-world application, particularly for automated warehousing management systems with multi-depth. As a learning-based and model-free method, it can autonomously learn proper retrieval strategies from interactions with the environment, directly using raw warehouse features. In contrast, traditional scheduling rules need to be designed by experts, requiring substantial expertise and leading to labor-intensive work (Braune et al., 2022).

Furthermore, once trained, the proposed agent does not require re-training or parameter tuning when applied to new structurally similar instances, as demonstrated in our generalization experiments described in Sect. 5.3. In practice, this significantly reduces the deployment cost and complexity. In addition, the learned policy is lightweight at the inference time, requiring only a few milliseconds per retrieval decision as shown in Table 10. This real-time decision-making capability is particularly well-suited for dynamic and time-sensitive environments, where rapid adaptation to varying workloads and due dates is crucial.

Additionally, while meta-heuristics such as GA may perform well on specific instances, they require extensive computation for each case and must be re-run whenever the problem changes (Rajwar et al., 2023). In practice, item due dates vary constantly, and the same location rarely stores the same type of item over time. In such dynamic and evolving environments, case-specific optimization is impractical, highlighting the advantage of learning-based methods.

### Practical implementation

The proposed DRL-based retrieval strategy can be integrated into existing warehouse management systems (WMS) in a modular manner. As shown in Fig. 5, the DRL agent performs a retrieval process similar to conventional rule-based dispatching logic: producing one decision at each decision point, that is, selecting the next item to retrieve from all directly accessible items. Therefore, it can serve as a replacement for existing retrieval rules (e.g., EDD and STT) without altering the rest of the warehouse control logic.

In practice, the agent can be deployed as a standalone decision module or integrated into the WMS via a lightweight API. For every decision timestep, the WMS collects the current system observation, encodes it as a graph-based state, and sends it to the agent module. The agent then returns the selected retrieved item, which is forwarded to the shuttle to execute the operation.

## Conclusion

This work proposes a novel DRL framework focused on the retrieval problem in a multi-deep AVS/RS, with the objective of total tardiness minimization. Moreover, we consider a more realistic scenario, in which all items are under a heterogeneous configuration, each with an independent and distinct due date. This setup remains an open research challenge.

To capture the complex topology in this warehouse, we first introduce a graph-based state representation that effectively captures both the warehouse's intricate topological structure and the features of the items. Specifically, each node represents the attributes of a corresponding storage location, whereas the edges encode structural relationships within the warehouse. To extract meaningful representations, we employ GraphSAGE, a GNN variant, to aggregate the features of items within each lane into the foremost items that are directly accessible for retrieval.

Furthermore, to integrate the local information extracted by graph-based state representation, we develop a innovative neural network architecture that merges a Transformer model with GraphSAGE. The Transformer model applies the attention mechanism across all directly accessible items throughout the warehouse to compute their priorities. This hybrid architecture effectively processes complex dependencies in multi-deep storage systems, ensuring both efficiency and strong generalizability.

This network is trained by PPO algorithm, which relies on an Actor-Critic structure. In the Actor network, the calculated priorities are transformed into selection probabilities using the SoftMax function, while the Critic outputs a scalar estimate of the current state's value. Additionally, a dense reward function is elaborated to facilitate the training process.

The proposed approach's effectiveness is substantiated by the experimental results. The developed DRL agent achieves a smooth training process and converges to a policy that significantly reduces total tardiness. In contrast, two comparative DRL agents in the ablation study using only the GNN model and the other using only the Transformer model, both struggle to converge to an effective strategy under the same hyperparameter configuration.

To validate the trained agent's generalization capability, we apply it to 128 diverse instances without retraining, with item counts ranging from 12 to 120 and due dates generated under four different settings. For comparison, we also adopt two heuristic rules, EDD and STT, to solve these instances. Across all the validation cases, the proposed approach achieves lower mean total tardiness in 119 instances, demonstrating its strong adaptability and superior performance across varying conditions.

Multiple promising directions for future research are possible building on our work. For example, integrating storage and relocation processes can further enhance warehouse management efficiency. These additional processes would introduce more environmental dynamics, such as fluctuating I/O points availability, lane occupancy, and unforeseen item arrival, which are difficult to handle using static methods. In contrast, the interactive decision-making nature of DRL inherently enables it to cope to real time dynamics, making it a promising approach for managing more complex and evolving warehouse operations. Additionally, although this study adopts a fixed warehouse layout, an interesting extension would be developing a more generalized model that supports diverse storage configurations, thereby improving real-world applicability. Although the proposed agent demonstrates promising generalization capability across different due-date settings and instance sizes, the performance superiority is less significant on small-scale instances. This suggests that the current generalization capability is partially dependent on the layout of the training instance. Therefore, training the agent on a mixture of warehouse layouts may further enhance its robustness and adaptability to a wider range of practical scenarios.

Addressing these aspects would not only optimize warehouse operations but also expand the applicability of multi-deep storage systems in dynamic and complex environments.

## Appendix A Pseudo-code of the Neural Networks in the Proposed DRL agent

**Input:** Graph $G = (\mathcal{V}, \mathcal{E})$; node features $x_i, \forall i \in \mathcal{V}$
**Output:** Node representation $z_i, \forall i \in \mathcal{V}$
**Parameters:** number of GNN layers $L$; GNN parameters $W^l$, $l \in \{1, \ldots L\}$; activation function $\sigma$; aggregation function $\mathrm{AGG}(\cdot)$; neighborhood $\mathcal{N}$

1: $h_i^0 \leftarrow x_i, \forall i \in \mathcal{V}$
2: **for** $l = 1 \ldots L$ **do**
3:     **for** each node $i \in \mathcal{V}$ **do**
4:         $h_{\mathcal{N}(i)}^l \leftarrow \mathrm{AGG}_l\big(\{\, h_u^{l-1}, \forall u \in \mathcal{N}(i)\}\big)$
5:         $h_i^l \leftarrow \sigma\big(W^l \cdot \mathrm{CONCAT}(h_i^{l-1}, h_{\mathcal{N}(i)}^l)\big)$
6:     **end for**
7:     Optionally normalize $h_i^l \leftarrow h_i^l / \left\|h_i^l\right\|_2, \forall i \in \mathcal{V}$
8: **end for**
9: $z_i \leftarrow h_i^L, \forall i \in \mathcal{V}$

**Algorithm 3** Graph processing mechanism in the GNN

**Input:** number of all directly accessible nodes $n$; GNN-embedded features of all directly accessible nodes $z_i, i \in \{1 \ldots n\}$
**Output:** Priority of all directly accessible node $p_i, i \in \{1 \ldots n\}$
**Parameters:** Query network parameters $W^Q$; Key network parameters $W^K$; Value network parameters $W^V$; Output dimension of Key network $d_k$

1: **for all** $i = 1, \ldots, n$ **do**
2:     Calculation of $q$ vector $q_i = z_i W^Q$
3:     Calculation of $k$ vector $k_i = z_i W^K$
4:     Calculation of $v$ vector $v_i = z_i W^V$
5:     Scaled Dot-Product Attention $y_i = \sum_{j=1}^{n} \frac{\exp\left(\frac{q_i \cdot k_j}{\sqrt{d_k}}\right)}{\sum_{j'=1}^{n} \exp\left(\frac{q_i \cdot k_{j'}}{\sqrt{d_k}}\right)} v_j$
6:     Calculation of priority $p_i = y_i W^{Prio}$
7: **end for**

**Algorithm 4** Attention mechanism and priority calculation algorithm in the Transformer model

## Appendix B Generalization capability of GNN-Only Agent and Transformer-Only Agent

**Table 11** Mean and minimum total tardiness obtained by the GNN-Transformer Agent, GNN-Only Agent, and Transformer-Only Agent in validation instances with due date configurations of $r = 0.125$ and $R = 0.75$

| | | | | GNN-Transformer Agent | | GNN-Only Agent | | Transformer-Only Agent | |
|---|---|---|---|---|---|---|---|---|---|
| $D_L$ | $N_A$ | $N_L$ | $I_i$ | Mean | Min | Mean | Min | Mean | Min |
| 2 | 1 | 6 | 1 | **23.98** | 0.00 | 79.71 | 27.30 | 48.91 | 0.00 |
| | | 8 | 2 | **52.67** | 9.44 | 196.51 | 60.93 | 125.11 | 77.22 |
| | | 10 | 3 | **62.27** | 4.40 | 265.17 | 112.50 | 156.28 | 95.55 |
| | | 12 | 4 | **68.57** | 0.00 | 419.67 | 169.60 | 237.02 | 53.16 |
| | 2 | 6 | 5 | **96.31** | 0.00 | 447.87 | 146.85 | 290.08 | 91.62 |
| | | 8 | 6 | **156.34** | 0.00 | 1024.22 | 792.70 | 626.26 | 314.83 |
| | | 10 | 7 | **192.81** | 0.00 | 1463.15 | 1153.46 | 915.18 | 735.66 |
| | | 12 | 8 | **381.38** | 0.00 | 2448.23 | 1658.51 | 1395.29 | 772.60 |
| 3 | 1 | 6 | 9 | **29.48** | 0.00 | 153.99 | 40.87 | 124.08 | 21.02 |
| | | 8 | 10 | **54.56** | 0.00 | 381.17 | 135.42 | 249.58 | 94.32 |
| | | 10 | 11 | **35.11** | 0.00 | 541.11 | 303.60 | 399.27 | 222.97 |
| | | 12 | 12 | **51.19** | 0.00 | 811.29 | 546.12 | 637.58 | 297.06 |
| | 2 | 6 | 13 | **73.05** | 0.00 | 898.97 | 577.56 | 897.86 | 521.32 |
| | | 8 | 14 | **160.32** | 0.00 | 2106.65 | 1462.77 | 1732.12 | 1216.74 |
| | | 10 | 15 | **232.70** | 30.83 | 3162.42 | 2245.04 | 2231.83 | 1663.73 |
| | | 12 | 16 | **300.67** | 19.15 | 5607.69 | 4727.00 | 4121.50 | 2755.99 |
| 4 | 1 | 6 | 17 | **46.85** | 0.00 | 267.57 | 62.31 | 203.40 | 32.13 |
| | | 8 | 18 | **78.97** | 0.00 | 466.43 | 131.91 | 529.72 | 111.98 |
| | | 10 | 19 | **76.30** | 0.00 | 759.46 | 315.17 | 756.96 | 434.07 |
| | | 12 | 20 | **144.94** | 0.00 | 1201.37 | 421.33 | 1241.03 | 774.79 |
| | 2 | 6 | 21 | **82.98** | 0.00 | 2153.22 | 1423.26 | 1842.28 | 750.52 |
| | | 8 | 22 | **235.61** | 0.00 | 5398.66 | 4066.01 | 3505.36 | 2248.79 |
| | | 10 | 23 | **190.17** | 0.00 | 7989.80 | 5575.34 | 5394.75 | 2775.96 |
| | | 12 | 24 | **466.13** | 0.00 | 13520.06 | 10074.92 | 8806.67 | 5380.51 |
| 5 | 1 | 6 | 25 | **84.63** | 0.00 | 540.80 | 188.85 | 388.32 | 84.08 |
| | | 8 | 26 | **237.55** | 35.12 | 1034.38 | 748.03 | 1087.12 | 574.26 |
| | | 10 | 27 | **328.09** | 19.88 | 1761.69 | 1197.75 | 1520.94 | 797.87 |
| | | 12 | 28 | **486.74** | 223.43 | 2891.13 | 2169.57 | 1992.67 | 1181.63 |
| | 2 | 6 | 29 | **309.53** | 0.00 | 5440.07 | 2704.90 | 3395.67 | 1797.05 |
| | | 8 | 30 | **611.00** | 41.11 | 12164.82 | 7439.93 | 6992.95 | 4256.33 |
| | | 10 | 31 | **849.24** | 119.12 | 18159.88 | 13345.52 | 9962.61 | 4772.96 |
| | | 12 | 32 | **1378.58** | 237.36 | 28333.49 | 20639.32 | 15145.00 | 11546.94 |

**Table 12** Mean and minimum total tardiness obtained by the GNN-Transformer Agent, GNN-Only Agent, and Transformer-Only Agent in validation instances with due date configurations of $r = 0.125$ and $R = 1.00$

| | | | | GNN-Transformer Agent | | GNN-Only Agent | | Transformer-Only Agent | |
|---|---|---|---|---|---|---|---|---|---|
| $D_L$ | $N_A$ | $N_L$ | $I_i$ | Mean | Min | Mean | Min | Mean | Min |
| 2 | 1 | 6 | 1 | **32.18** | 0.00 | 105.29 | 33.95 | 74.34 | 32.70 |
| | | 8 | 2 | **71.07** | 8.59 | 253.21 | 88.61 | 162.68 | 89.90 |
| | | 10 | 3 | **71.81** | 3.05 | 377.74 | 189.57 | 238.21 | 146.56 |
| | | 12 | 4 | **95.04** | 0.00 | 579.31 | 256.28 | 316.46 | 148.69 |
| | 2 | 6 | 5 | **121.99** | 0.00 | 500.94 | 112.34 | 350.67 | 126.22 |
| | | 8 | 6 | **197.09** | 0.00 | 1149.18 | 847.63 | 811.73 | 389.44 |
| | | 10 | 7 | **278.30** | 0.00 | 1728.94 | 1327.82 | 1188.66 | 736.55 |
| | | 12 | 8 | **505.58** | 22.25 | 2850.86 | 1990.90 | 1603.96 | 880.56 |
| 3 | 1 | 6 | 9 | **56.45** | 0.00 | 221.45 | 55.58 | 190.98 | 0.00 |
| | | 8 | 10 | **115.57** | 0.00 | 516.54 | 204.55 | 350.02 | 134.09 |
| | | 10 | 11 | **94.92** | 0.00 | 759.56 | 468.80 | 523.18 | 212.75 |
| | | 12 | 12 | **129.83** | 0.00 | 1116.37 | 793.59 | 833.08 | 416.08 |
| | 2 | 6 | 13 | **173.84** | 0.00 | 1183.18 | 752.44 | 1168.10 | 683.43 |
| | | 8 | 14 | **303.05** | 0.00 | 2416.35 | 1603.53 | 2174.12 | 1543.99 |
| | | 10 | 15 | **491.76** | 81.61 | 3781.19 | 2386.96 | 2911.30 | 2408.41 |
| | | 12 | 16 | **647.07** | 170.97 | 6553.12 | 5707.42 | 5119.38 | 3413.99 |
| 4 | 1 | 6 | 17 | **88.04** | 0.00 | 384.79 | 82.48 | 275.92 | 0.00 |
| | | 8 | 18 | **152.54** | 0.00 | 705.78 | 350.08 | 709.44 | 167.98 |
| | | 10 | 19 | **209.24** | 0.00 | 1154.98 | 492.39 | 1128.38 | 686.86 |
| | | 12 | 20 | **323.45** | 0.00 | 1630.99 | 613.24 | 1534.94 | 1048.94 |
| | 2 | 6 | 21 | **261.96** | 0.00 | 2611.47 | 1642.23 | 2446.68 | 1351.45 |
| | | 8 | 22 | **568.62** | 74.68 | 5759.57 | 4055.16 | 4274.20 | 3236.00 |
| | | 10 | 23 | **733.75** | 59.42 | 9283.77 | 5851.17 | 7096.63 | 4385.84 |
| | | 12 | 24 | **1261.91** | 6.22 | 15111.83 | 10559.26 | 11103.51 | 7542.72 |
| 5 | 1 | 6 | 25 | **154.23** | 0.00 | 798.34 | 273.54 | 574.08 | 161.39 |
| | | 8 | 26 | **374.06** | 37.16 | 1499.71 | 1112.76 | 1536.87 | 1102.69 |
| | | 10 | 27 | **577.48** | 11.12 | 2423.84 | 1612.70 | 2067.13 | 892.71 |
| | | 12 | 28 | **903.21** | 326.26 | 3984.29 | 2884.99 | 2996.96 | 2038.29 |
| | 2 | 6 | 29 | **737.12** | 7.00 | 6310.35 | 2940.64 | 4376.79 | 2481.70 |
| | | 8 | 30 | **1504.34** | 518.82 | 13406.46 | 8086.21 | 8981.95 | 5992.94 |
| | | 10 | 31 | **1768.82** | 586.30 | 20399.79 | 14423.30 | 12456.62 | 8522.85 |
| | | 12 | 32 | **2986.30** | 1142.40 | 31082.18 | 22502.53 | 19424.42 | 15027.28 |

**Table 13** Mean and minimum total tardiness obtained by the GNN-Transformer Agent, GNN-Only Agent, and Transformer-Only Agent in validation instances with due date configurations of $r = 0.25$ and $R = 0.75$

| | | | | GNN-Transformer Agent | | GNN-Only Agent | | Transformer-Only Agent | |
|---|---|---|---|---|---|---|---|---|---|
| $D_L$ | $N_A$ | $N_L$ | $I_i$ | Mean | Min | Mean | Min | Mean | Min |
| 2 | 1 | 6 | 1 | **53.76** | 12.40 | 121.31 | 72.76 | 83.80 | 3.19 |
| | | 8 | 2 | **133.73** | 15.44 | 279.66 | 138.29 | 198.76 | 91.41 |
| | | 10 | 3 | **158.65** | 39.41 | 414.65 | 253.41 | 320.81 | 235.54 |
| | | 12 | 4 | **226.38** | 15.96 | 627.63 | 392.44 | 466.54 | 233.27 |
| | 2 | 6 | 5 | **235.53** | 15.49 | 718.71 | 430.21 | 547.48 | 286.00 |
| | | 8 | 6 | **439.70** | 184.18 | 1560.91 | 1136.77 | 1151.29 | 812.87 |
| | | 10 | 7 | **547.96** | 67.20 | 2412.99 | 2064.58 | 1575.48 | 1121.67 |
| | | 12 | 8 | **980.68** | 339.90 | 3935.35 | 3162.50 | 2526.19 | 1598.49 |
| 3 | 1 | 6 | 9 | **97.65** | 0.00 | 339.84 | 181.19 | 220.71 | 76.67 |
| | | 8 | 10 | **214.44** | 10.53 | 657.05 | 236.14 | 487.50 | 297.86 |
| | | 10 | 11 | **159.98** | 0.00 | 959.59 | 700.95 | 755.28 | 443.44 |
| | | 12 | 12 | **331.61** | 66.85 | 1603.55 | 1229.27 | 1227.83 | 748.25 |
| | 2 | 6 | 13 | **349.57** | 77.74 | 1871.68 | 1215.29 | 1491.85 | 1098.34 |
| | | 8 | 14 | **710.46** | 420.38 | 3836.29 | 2775.06 | 3033.47 | 2177.43 |
| | | 10 | 15 | **993.27** | 268.85 | 5894.52 | 4884.75 | 4275.88 | 3643.84 |
| | | 12 | 16 | **1603.11** | 943.51 | 9877.31 | 8833.37 | 6968.46 | 5367.80 |
| 4 | 1 | 6 | 17 | **165.77** | 0.00 | 579.92 | 267.52 | 377.98 | 31.40 |
| | | 8 | 18 | **287.36** | 25.70 | 1002.04 | 599.21 | 1014.67 | 459.26 |
| | | 10 | 19 | **354.45** | 18.30 | 1576.11 | 929.78 | 1573.45 | 1277.90 |
| | | 12 | 20 | **643.93** | 24.00 | 2373.63 | 1362.50 | 2344.08 | 1525.63 |
| | 2 | 6 | 21 | **513.82** | 0.00 | 4143.16 | 3130.68 | 3089.69 | 1879.72 |
| | | 8 | 22 | **1402.09** | 425.19 | 9159.21 | 7565.52 | 6402.89 | 5076.09 |
| | | 10 | 23 | **1658.24** | 367.79 | 14090.54 | 11420.93 | 9619.18 | 5956.60 |
| | | 12 | 24 | **2733.56** | 671.81 | 22241.80 | 18738.45 | 15030.47 | 11176.03 |
| 5 | 1 | 6 | 25 | **290.92** | 23.66 | 1102.49 | 562.13 | 861.06 | 400.02 |
| | | 8 | 26 | **682.10** | 216.41 | 2025.63 | 1492.77 | 1963.50 | 1121.88 |
| | | 10 | 27 | **995.90** | 402.55 | 3293.86 | 2704.48 | 2931.40 | 1831.15 |
| | | 12 | 28 | **1718.27** | 1054.19 | 5139.32 | 4299.87 | 4275.71 | 3303.83 |
| | 2 | 6 | 29 | **1376.10** | 585.96 | 9482.18 | 6790.63 | 6070.46 | 3480.62 |
| | | 8 | 30 | **2776.26** | 1626.59 | 19368.09 | 14738.61 | 12426.97 | 8489.95 |
| | | 10 | 31 | **3542.06** | 1892.70 | 29278.89 | 23182.22 | 17571.23 | 12811.30 |
| | | 12 | 32 | **6325.23** | 4086.11 | 42855.29 | 35474.84 | 27061.37 | 21576.86 |

**Table 14** Mean and minimum total tardiness obtained by the GNN-Transformer Agent, GNN-Only Agent, and Transformer-Only Agent in validation instances with due date configurations of $r = 0.25$ and $R = 1.00$

| | | | | GNN-Transformer Agent | | GNN-Only Agent | | Transformer-Only Agent | |
|---|---|---|---|---|---|---|---|---|---|
| $D_L$ | $N_A$ | $N_L$ | $I_i$ | Mean | Min | Mean | Min | Mean | Min |
| 2 | 1 | 6 | 1 | **61.35** | 11.98 | 161.64 | 91.02 | 115.10 | 38.35 |
| | | 8 | 2 | **140.79** | 28.59 | 365.57 | 174.74 | 251.31 | 137.65 |
| | | 10 | 3 | **173.40** | 35.55 | 529.94 | 365.72 | 395.68 | 183.05 |
| | | 12 | 4 | **250.60** | 7.28 | 797.69 | 456.81 | 521.77 | 276.36 |
| | 2 | 6 | 5 | **259.84** | 0.00 | 825.29 | 438.08 | 662.35 | 403.91 |
| | | 8 | 6 | **479.79** | 142.92 | 1707.33 | 1170.14 | 1261.25 | 792.71 |
| | | 10 | 7 | **602.21** | 72.37 | 2685.89 | 2211.28 | 2043.28 | 1444.35 |
| | | 12 | 8 | **1091.18** | 237.55 | 4228.86 | 3167.30 | 2824.05 | 1727.80 |
| 3 | 1 | 6 | 9 | **113.46** | 1.02 | 405.34 | 201.58 | 294.29 | 34.22 |
| | | 8 | 10 | **276.38** | 54.05 | 838.64 | 301.55 | 587.92 | 348.65 |
| | | 10 | 11 | **233.13** | 24.84 | 1257.45 | 884.93 | 916.23 | 362.59 |
| | | 12 | 12 | **485.78** | 122.77 | 1858.15 | 1384.17 | 1638.69 | 852.48 |
| | 2 | 6 | 13 | **488.61** | 0.00 | 1975.55 | 1383.82 | 1878.26 | 1174.75 |
| | | 8 | 14 | **886.83** | 142.68 | 4126.57 | 3005.68 | 3710.94 | 2863.98 |
| | | 10 | 15 | **1226.75** | 267.46 | 6293.85 | 4789.01 | 4962.58 | 4152.54 |
| | | 12 | 16 | **1938.80** | 1081.17 | 10523.80 | 9416.42 | 8131.72 | 6411.82 |
| 4 | 1 | 6 | 17 | **231.30** | 0.00 | 730.81 | 299.33 | 495.10 | 0.41 |
| | | 8 | 18 | **416.87** | 52.60 | 1282.11 | 753.23 | 1246.89 | 384.31 |
| | | 10 | 19 | **555.44** | 26.34 | 2013.27 | 1182.08 | 2019.52 | 1611.67 |
| | | 12 | 20 | **929.32** | 56.63 | 3106.15 | 1805.53 | 2844.11 | 1999.54 |
| | 2 | 6 | 21 | **858.27** | 64.48 | 4371.97 | 3179.58 | 3759.49 | 2725.78 |
| | | 8 | 22 | **1768.82** | 454.02 | 9466.55 | 7649.84 | 7271.19 | 5956.23 |
| | | 10 | 23 | **2570.29** | 932.45 | 14750.86 | 11020.75 | 11942.57 | 9289.58 |
| | | 12 | 24 | **4211.94** | 1103.05 | 23885.27 | 19413.41 | 17946.84 | 13847.59 |
| 5 | 1 | 6 | 25 | **447.08** | 25.64 | 1382.50 | 746.77 | 1066.48 | 391.10 |
| | | 8 | 26 | **893.43** | 454.56 | 2471.00 | 1892.48 | 2515.01 | 1990.62 |
| | | 10 | 27 | **1393.92** | 566.85 | 4196.07 | 3347.69 | 3590.61 | 2137.95 |
| | | 12 | 28 | **2265.49** | 1117.73 | 6603.03 | 5489.22 | 5170.12 | 3803.62 |
| | 2 | 6 | 29 | **2111.50** | 935.21 | 9942.51 | 6474.08 | 7413.56 | 4424.19 |
| | | 8 | 30 | **3945.59** | 2361.35 | 20047.15 | 14785.04 | 14933.23 | 12008.09 |
| | | 10 | 31 | **5086.61** | 1943.47 | 31097.32 | 23953.98 | 21532.24 | 16944.78 |
| | | 12 | 32 | **8507.96** | 5479.60 | 46369.89 | 37454.93 | 32033.07 | 23759.86 |

## Appendix C Comparison with additional rules

**Table 15** Mean total tardiness obtained by the GNN-Transformer Agent, STT, EDD, LST and Random in validation instances with due date configurations of $r = 0.125$ and $R = 0.75$

| $D_L$ | $N_A$ | $N_L$ | $I_i$ | GNN-Transformer Agent | STT | EDD | LST | Random |
|---|---|---|---|---|---|---|---|---|
| 2 | 1 | 6 | 1 | 23.98 | 36.12 | **22.82** | 23.74 | 75.41 |
| | | 8 | 2 | **52.67** | 100.46 | 65.87 | 69.42 | 203.17 |
| | | 10 | 3 | **62.27** | 144.59 | 83.81 | 85.54 | 307.01 |
| | | 12 | 4 | **68.57** | 172.35 | 118.29 | 121.50 | 597.60 |
| | 2 | 6 | 5 | 96.31 | 157.74 | **94.87** | 103.65 | 881.26 |
| | | 8 | 6 | **156.34** | 381.44 | 227.95 | 287.46 | 1759.96 |
| | | 10 | 7 | **192.81** | 468.27 | 331.10 | 360.14 | 2861.97 |
| | | 12 | 8 | **381.38** | 675.05 | 577.69 | 723.63 | 4704.88 |
| 3 | 1 | 6 | 9 | **29.48** | 101.46 | 82.47 | 83.84 | 177.31 |
| | | 8 | 10 | **54.56** | 278.12 | 209.98 | 218.09 | 509.69 |
| | | 10 | 11 | **35.11** | 274.30 | 277.86 | 290.05 | 744.20 |
| | | 12 | 12 | **51.19** | 532.46 | 498.21 | 508.23 | 1218.07 |
| | 2 | 6 | 13 | **73.05** | 425.81 | 454.67 | 505.51 | 2542.90 |
| | | 8 | 14 | **160.32** | 1003.49 | 1097.44 | 1182.14 | 5137.39 |
| | | 10 | 15 | **232.70** | 1404.18 | 1615.25 | 1799.41 | 8199.51 |
| | | 12 | 16 | **300.67** | 2167.20 | 2418.40 | 2615.11 | 13739.13 |
| 4 | 1 | 6 | 17 | **46.85** | 213.68 | 218.40 | 218.40 | 279.81 |
| | | 8 | 18 | **78.97** | 393.25 | 398.95 | 407.76 | 831.79 |
| | | 10 | 19 | **76.30** | 752.90 | 651.37 | 653.88 | 1109.36 |
| | | 12 | 20 | **144.94** | 994.21 | 981.05 | 992.62 | 2395.91 |
| | 2 | 6 | 21 | **82.98** | 798.49 | 1274.28 | 1382.04 | 5245.89 |
| | | 8 | 22 | **235.61** | 1750.93 | 2797.87 | 2989.50 | 11602.44 |
| | | 10 | 23 | **190.17** | 2893.25 | 4246.14 | 4609.21 | 18434.95 |
| | | 12 | 24 | **466.13** | 4668.36 | 6398.88 | 6926.45 | 28535.13 |
| 5 | 1 | 6 | 25 | **84.63** | 353.15 | 332.48 | 333.98 | 586.72 |
| | | 8 | 26 | **237.55** | 934.00 | 794.77 | 800.97 | 1322.92 |
| | | 10 | 27 | **328.09** | 1379.14 | 1292.30 | 1323.30 | 2362.26 |
| | | 12 | 28 | **486.74** | 1844.85 | 2053.73 | 2118.10 | 3907.24 |
| | 2 | 6 | 29 | **309.53** | 1907.28 | 2681.49 | 2994.16 | 10884.11 |
| | | 8 | 30 | **611.00** | 3203.46 | 5196.69 | 5562.20 | 21945.60 |
| | | 10 | 31 | **849.24** | 5135.56 | 7378.59 | 7943.05 | 33121.89 |
| | | 12 | 32 | **1378.58** | 7913.43 | 11828.96 | 12624.57 | 49291.62 |

**Table 16** Mean total tardiness obtained by the GNN-Transformer Agent, STT, EDD, LST and Random in validation instances with due date configurations of $r = 0.125$ and $R = 1.00$

| $D_L$ | $N_A$ | $N_L$ | $I_i$ | GNN-Transformer Agent | STT | EDD | LST | Random |
|---|---|---|---|---|---|---|---|---|
| 2 | 1 | 6 | 1 | **32.18** | 50.50 | 32.86 | 33.81 | 98.98 |
| | | 8 | 2 | **71.07** | 139.20 | 86.71 | 90.60 | 256.50 |
| | | 10 | 3 | **71.81** | 205.46 | 110.81 | 110.61 | 370.92 |
| | | 12 | 4 | **95.04** | 254.95 | 153.56 | 158.56 | 706.54 |
| | 2 | 6 | 5 | **121.99** | 260.72 | 127.29 | 134.69 | 980.59 |
| | | 8 | 6 | **197.09** | 535.88 | 235.44 | 258.95 | 1867.56 |
| | | 10 | 7 | **278.30** | 706.42 | 399.28 | 423.01 | 3058.16 |
| | | 12 | 8 | **505.58** | 989.91 | 649.02 | 722.80 | 4958.33 |
| 3 | 1 | 6 | 9 | **56.45** | 152.32 | 116.24 | 118.14 | 233.33 |
| | | 8 | 10 | **115.57** | 389.69 | 284.63 | 295.66 | 635.53 |
| | | 10 | 11 | **94.92** | 430.14 | 396.30 | 404.91 | 941.31 |
| | | 12 | 12 | **129.83** | 784.24 | 671.17 | 679.83 | 1491.28 |
| | 2 | 6 | 13 | **173.84** | 664.56 | 593.46 | 632.29 | 2774.66 |
| | | 8 | 14 | **303.05** | 1456.96 | 1379.21 | 1442.11 | 5511.66 |
| | | 10 | 15 | **491.76** | 2164.97 | 2078.76 | 2234.63 | 8902.31 |
| | | 12 | 16 | **647.07** | 3237.23 | 3062.98 | 3198.38 | 14749.15 |
| 4 | 1 | 6 | 17 | **88.04** | 323.49 | 310.57 | 310.57 | 412.81 |
| | | 8 | 18 | **152.54** | 604.82 | 565.87 | 562.45 | 1072.04 |
| | | 10 | 19 | **209.24** | 1107.78 | 919.82 | 918.04 | 1500.21 |
| | | 12 | 20 | **323.45** | 1491.61 | 1380.44 | 1387.87 | 3032.75 |
| | 2 | 6 | 21 | **261.96** | 1249.46 | 1666.74 | 1781.84 | 5695.57 |
| | | 8 | 22 | **568.62** | 2585.01 | 3497.88 | 3661.22 | 12493.79 |
| | | 10 | 23 | **733.75** | 4422.63 | 5412.51 | 5680.08 | 19851.63 |
| | | 12 | 24 | **1261.91** | 6844.04 | 8069.44 | 8516.04 | 30460.39 |
| 5 | 1 | 6 | 25 | **154.23** | 536.03 | 474.24 | 471.88 | 796.91 |
| | | 8 | 26 | **374.06** | 1332.52 | 1071.88 | 1075.48 | 1754.95 |
| | | 10 | 27 | **577.48** | 2008.19 | 1799.62 | 1825.36 | 3050.74 |
| | | 12 | 28 | **903.21** | 2676.18 | 2812.19 | 2836.01 | 5002.64 |
| | 2 | 6 | 29 | **737.12** | 2889.68 | 3422.35 | 3657.15 | 11813.04 |
| | | 8 | 30 | **1504.34** | 4839.32 | 6550.51 | 6743.43 | 23708.18 |
| | | 10 | 31 | **1768.82** | 7810.25 | 9328.38 | 9748.28 | 35734.88 |
| | | 12 | 32 | **2986.30** | 11745.48 | 14727.72 | 15362.79 | 52682.35 |

**Table 17** Mean total tardiness obtained by the GNN-Transformer Agent, STT, EDD, LST and Random in validation instances with due date configurations of $r = 0.25$ and $R = 0.75$

| $D_L$ | $N_A$ | $N_L$ | $I_i$ | GNN-Transformer Agent | STT | EDD | LST | Random |
|---|---|---|---|---|---|---|---|---|
| 2 | 1 | 6 | 1 | 53.76 | 73.45 | **43.06** | 46.72 | 128.38 |
| | | 8 | 2 | 133.73 | 191.47 | **112.74** | 120.19 | 338.53 |
| | | 10 | 3 | **158.65** | 273.21 | 160.83 | 168.18 | 504.32 |
| | | 12 | 4 | **226.38** | 340.53 | 229.62 | 247.68 | 939.76 |
| | 2 | 6 | 5 | 235.53 | 353.96 | **215.33** | 241.40 | 1223.32 |
| | | 8 | 6 | **439.70** | 749.71 | 548.57 | 687.01 | 2453.51 |
| | | 10 | 7 | **547.96** | 948.97 | 750.80 | 857.79 | 4001.88 |
| | | 12 | 8 | **980.68** | 1367.19 | 1312.01 | 1601.00 | 6472.73 |
| 3 | 1 | 6 | 9 | **97.65** | 218.06 | 163.89 | 169.69 | 302.07 |
| | | 8 | 10 | **214.44** | 508.31 | 368.08 | 379.15 | 823.72 |
| | | 10 | 11 | **159.98** | 614.13 | 536.88 | 547.53 | 1231.21 |
| | | 12 | 12 | **331.61** | 1047.24 | 872.88 | 889.11 | 1998.50 |
| | 2 | 6 | 13 | **349.57** | 908.83 | 856.51 | 970.67 | 3606.24 |
| | | 8 | 14 | **710.46** | 1938.07 | 2012.15 | 2200.25 | 7195.70 |
| | | 10 | 15 | **993.27** | 2852.73 | 2897.55 | 3254.96 | 11382.64 |
| | | 12 | 16 | **1603.11** | 4340.82 | 4613.47 | 5012.07 | 18614.42 |
| 4 | 1 | 6 | 17 | **165.77** | 419.68 | 403.05 | 403.05 | 588.60 |
| | | 8 | 18 | **287.36** | 806.58 | 726.26 | 736.58 | 1457.81 |
| | | 10 | 19 | **354.45** | 1432.50 | 1170.01 | 1179.30 | 2162.01 |
| | | 12 | 20 | **643.93** | 2031.57 | 1799.79 | 1799.90 | 4017.81 |
| | 2 | 6 | 21 | **513.82** | 1761.38 | 2243.75 | 2428.80 | 7546.43 |
| | | 8 | 22 | **1402.09** | 3456.87 | 4729.55 | 5115.74 | 15737.04 |
| | | 10 | 23 | **1658.24** | 5736.38 | 7194.92 | 7785.04 | 25338.87 |
| | | 12 | 24 | **2733.56** | 9057.90 | 10986.70 | 11921.37 | 38938.86 |
| 5 | 1 | 6 | 25 | **290.92** | 767.71 | 645.07 | 652.80 | 1063.52 |
| | | 8 | 26 | **682.10** | 1751.33 | 1418.15 | 1434.78 | 2374.80 |
| | | 10 | 27 | **995.90** | 2631.66 | 2299.64 | 2352.77 | 4056.48 |
| | | 12 | 28 | **1718.27** | 3679.90 | 3644.36 | 3729.11 | 6581.49 |
| | 2 | 6 | 29 | **1376.10** | 3849.12 | 4547.84 | 4964.68 | 15203.13 |
| | | 8 | 30 | **2776.26** | 6515.93 | 8871.46 | 9417.15 | 29856.98 |
| | | 10 | 31 | **3542.06** | 10264.59 | 12353.86 | 13277.45 | 45555.49 |
| | | 12 | 32 | **6325.23** | 15401.23 | 20117.73 | 21421.82 | 67768.82 |

**Table 18** Mean total tardiness obtained by the GNN-Transformer Agent, STT, EDD, LST and Random in validation instances with due date configurations of $r = 0.25$ and $R = 1.00$

| $D_L$ | $N_A$ | $N_L$ | $I_i$ | GNN-Transformer Agent | STT | EDD | LST | Random |
|---|---|---|---|---|---|---|---|---|
| 2 | 1 | 6 | 1 | 61.35 | 92.93 | **55.31** | 59.46 | 153.52 |
| | | 8 | 2 | 140.79 | 240.56 | **135.04** | 141.61 | 391.07 |
| | | 10 | 3 | **173.40** | 342.59 | 189.04 | 192.69 | 576.53 |
| | | 12 | 4 | **250.60** | 443.70 | 262.33 | 269.55 | 1043.37 |
| | 2 | 6 | 5 | 259.84 | 464.69 | **211.94** | 227.57 | 1331.02 |
| | | 8 | 6 | 479.79 | 935.84 | **447.18** | 523.17 | 2557.31 |
| | | 10 | 7 | **602.21** | 1244.97 | 688.22 | 735.01 | 4201.21 |
| | | 12 | 8 | **1091.18** | 1737.16 | 1119.57 | 1306.61 | 6746.63 |
| 3 | 1 | 6 | 9 | **113.46** | 281.74 | 208.75 | 214.55 | 373.25 |
| | | 8 | 10 | **276.38** | 645.74 | 454.73 | 461.39 | 956.16 |
| | | 10 | 11 | **233.13** | 799.87 | 671.61 | 688.38 | 1430.32 |
| | | 12 | 12 | **485.78** | 1345.75 | 1073.48 | 1082.09 | 2268.88 |
| | 2 | 6 | 13 | **488.61** | 1217.04 | 1000.86 | 1070.81 | 3858.80 |
| | | 8 | 14 | **886.83** | 2493.01 | 2126.47 | 2220.05 | 7555.45 |
| | | 10 | 15 | **1226.75** | 3813.84 | 3297.04 | 3484.98 | 12121.53 |
| | | 12 | 16 | **1938.80** | 5655.42 | 4983.42 | 5199.78 | 19561.67 |
| 4 | 1 | 6 | 17 | **231.30** | 559.96 | 519.60 | 519.60 | 746.58 |
| | | 8 | 18 | **416.87** | 1050.67 | 924.14 | 927.04 | 1733.35 |
| | | 10 | 19 | **555.44** | 1880.38 | 1509.43 | 1520.03 | 2558.72 |
| | | 12 | 20 | **929.32** | 2708.00 | 2269.22 | 2277.69 | 4671.40 |
| | 2 | 6 | 21 | **858.27** | 2352.17 | 2677.33 | 2807.75 | 7977.11 |
| | | 8 | 22 | **1768.82** | 4547.89 | 5349.06 | 5588.04 | 16541.30 |
| | | 10 | 23 | **2570.29** | 7685.24 | 8337.83 | 8669.85 | 26745.32 |
| | | 12 | 24 | **4211.94** | 11876.39 | 12440.55 | 12996.04 | 40740.38 |
| 5 | 1 | 6 | 25 | **447.08** | 1014.40 | 812.24 | 818.31 | 1310.62 |
| | | 8 | 26 | **893.43** | 2218.92 | 1745.79 | 1755.07 | 2822.77 |
| | | 10 | 27 | **1393.92** | 3503.57 | 2869.00 | 2913.65 | 4894.24 |
| | | 12 | 28 | **2265.49** | 4887.47 | 4523.17 | 4553.33 | 7876.84 |
| | 2 | 6 | 29 | **2111.50** | 5046.06 | 5292.53 | 5578.30 | 16153.44 |
| | | 8 | 30 | **3945.59** | 8645.77 | 10072.31 | 10340.42 | 31700.85 |
| | | 10 | 31 | **5086.61** | 13666.75 | 14634.19 | 15251.09 | 48164.70 |
| | | 12 | 32 | **8507.96** | 20457.41 | 22897.87 | 23738.66 | 71283.75 |

## Appendix D The layouts of validation instances

We evaluate the generalization capability of the DRL agent that trained in Sec. 5 on 128 instances with 32 different layouts, where each layout generates four instances under four different due date configurations. These layouts are demonstrated in Figs. 19, 20, 21, 22, 23, 24, 25 and 26 in graph-based representation. The due dates of items are not represented by varying intensity of blue colors, since the due date of each item will be randomly regenerated for each run and Figs. 19, 20, 21, 22, 23, 24, 25 and 26 solely illustrate the warehouse layouts.

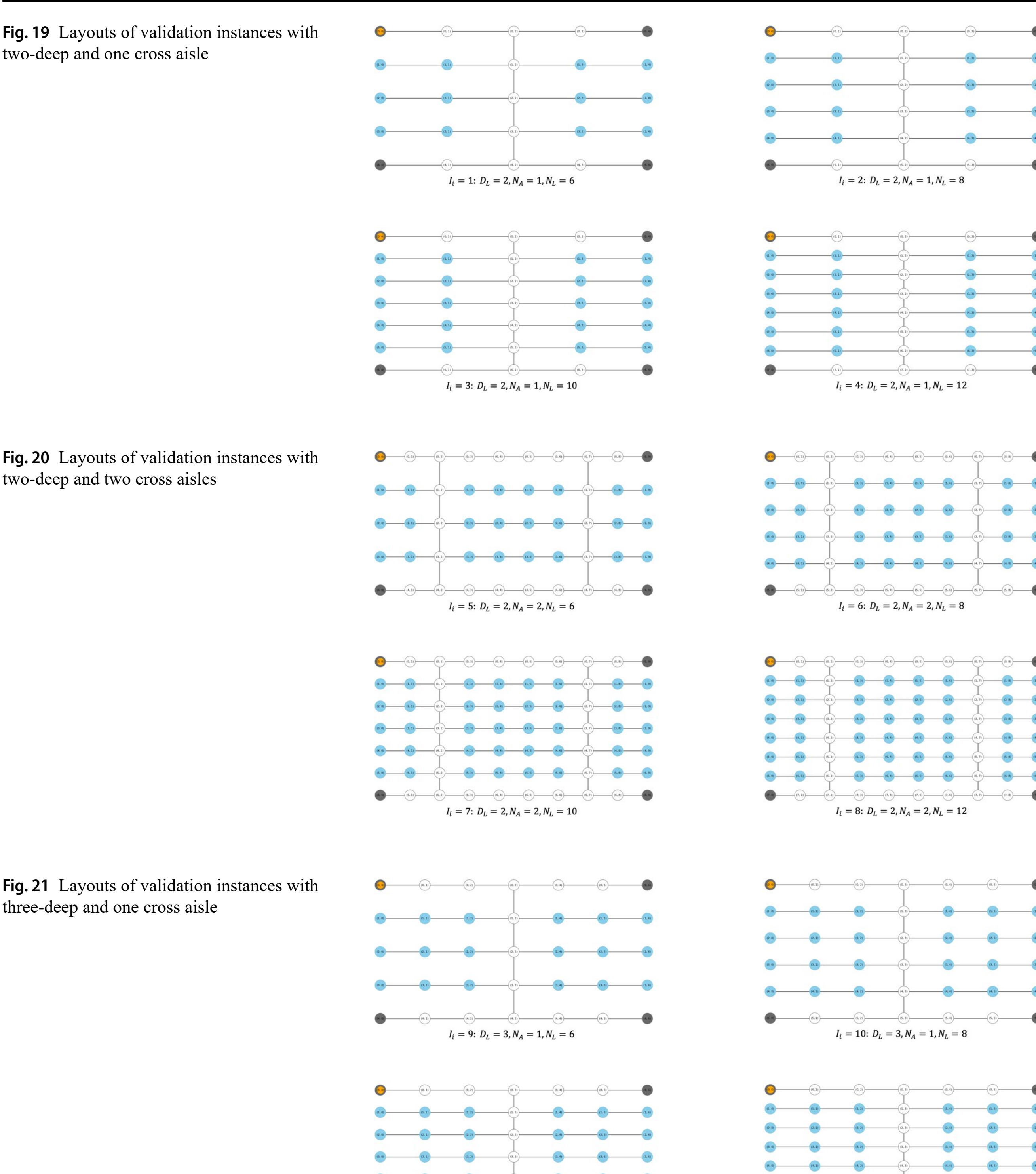

**Fig. 19** Layouts of validation instances with two-deep and one cross aisle

**Fig. 20** Layouts of validation instances with two-deep and two cross aisles

**Fig. 21** Layouts of validation instances with three-deep and one cross aisle

**Fig. 22** Layouts of validation instances with three-deep and two cross aisles

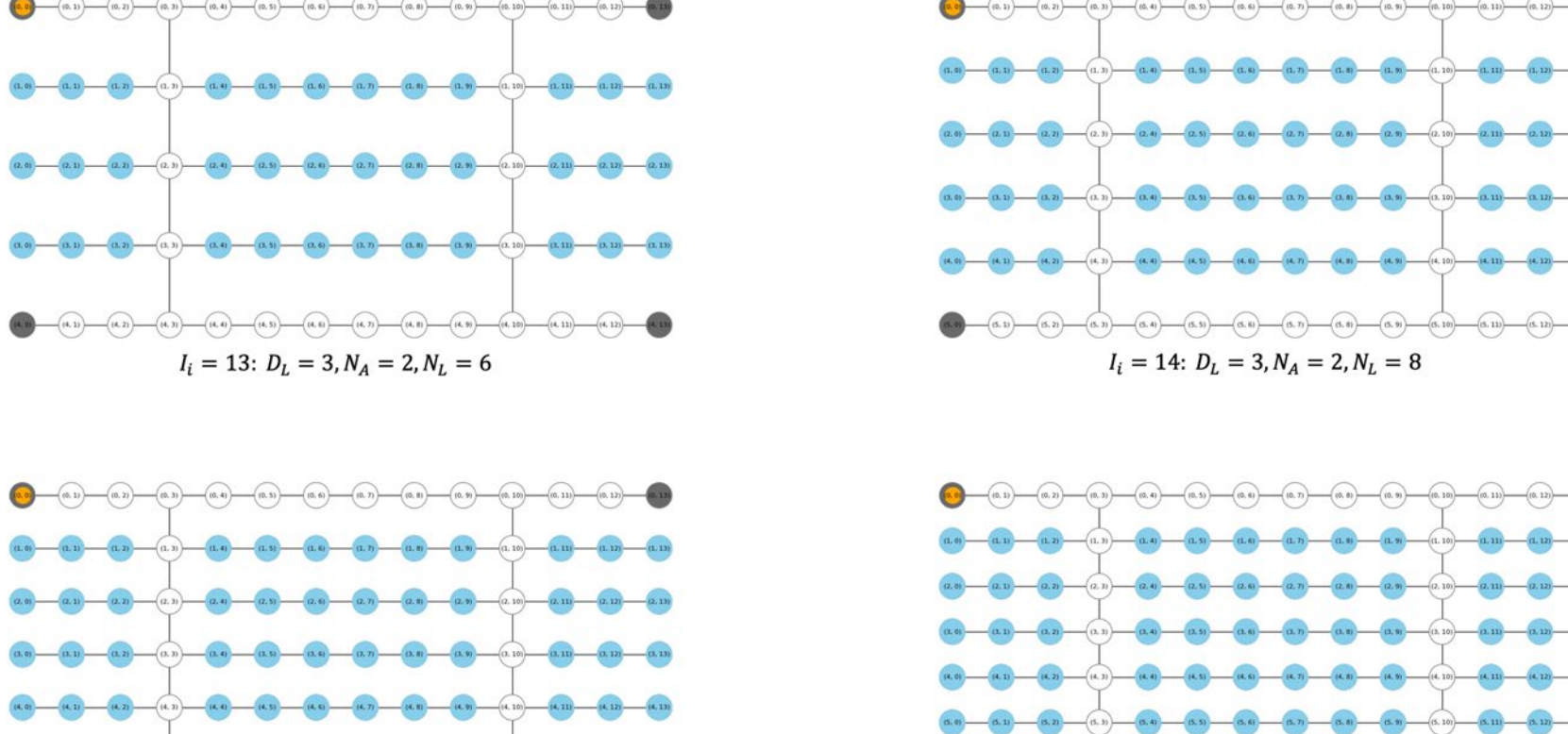

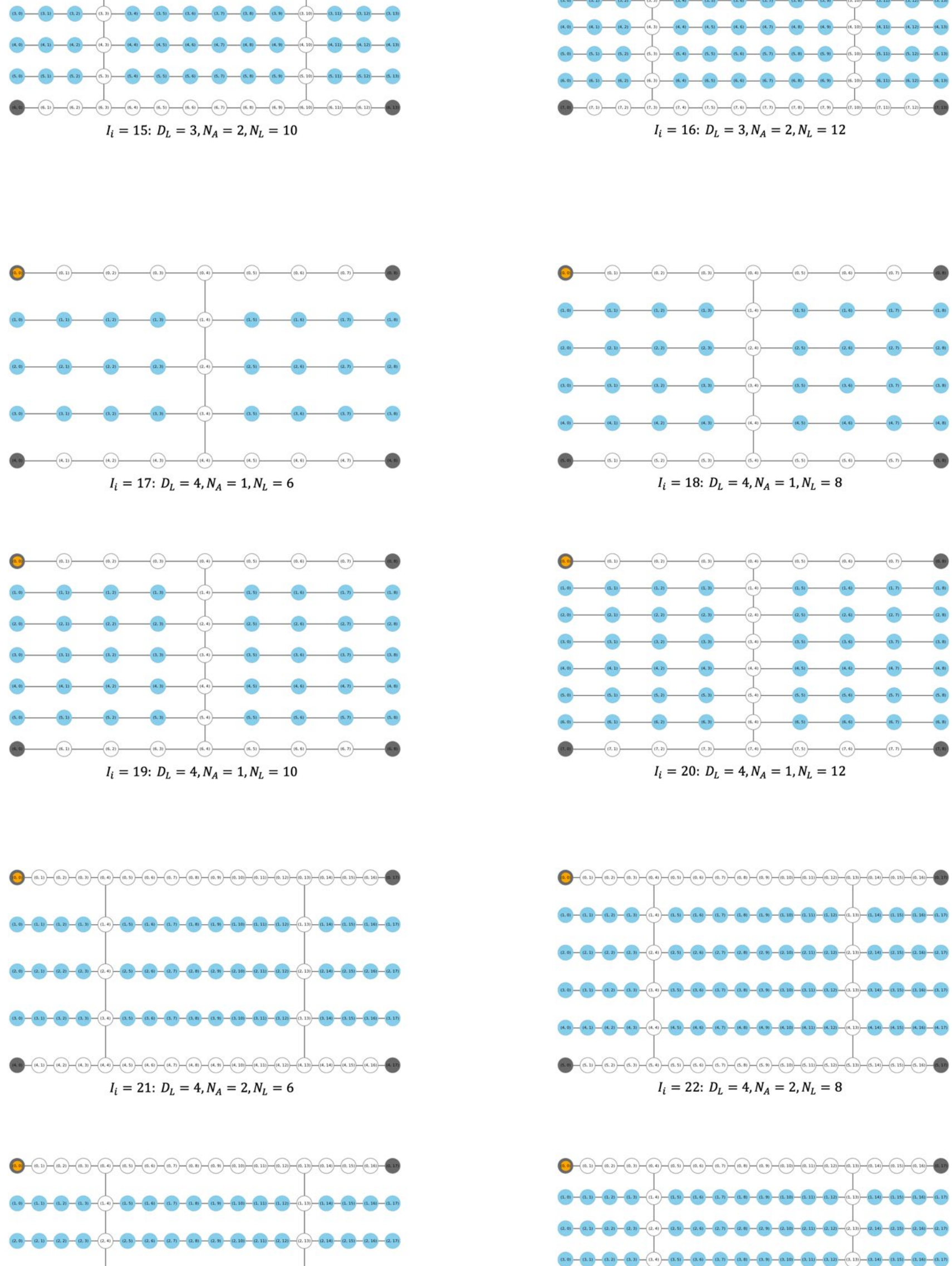

**Fig. 23** Layouts of validation instances with four-deep and one cross aisle

**Fig. 24** Layouts of validation instances with four-deep and two cross aisles

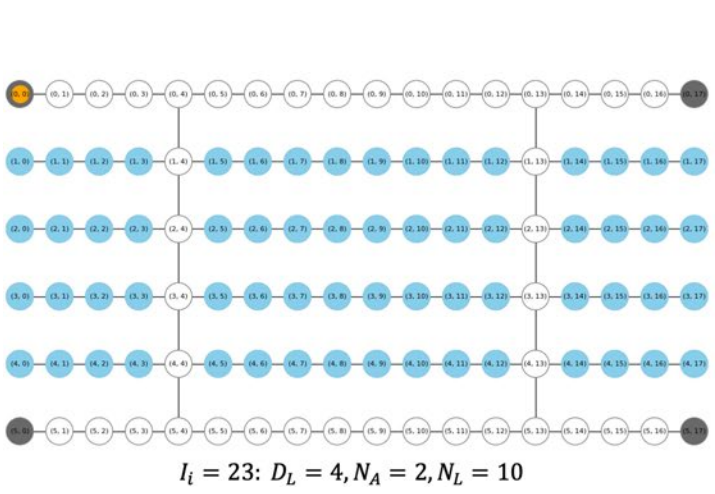

$l_i = 24$: $D_L = 4, N_A = 2, N_L = 12$

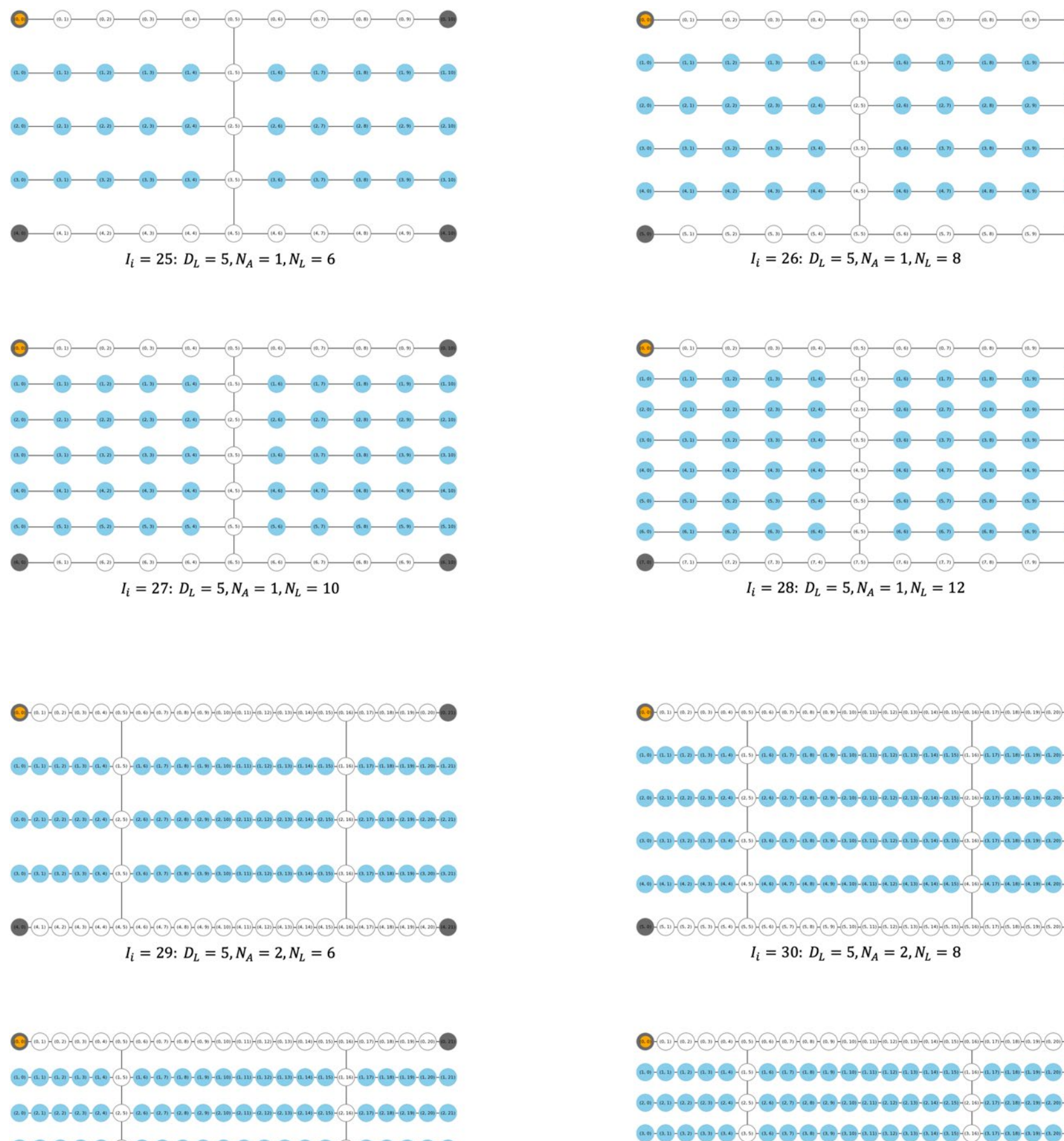

**Fig. 25** Layouts of validation instances with five-deep and one cross aisle

**Fig. 26** Layouts of validation instances with five-deep and two cross aisles

**Acknowledgements** The work of Funing Li, Ruben Noortwyck, Jifeng Zhou, and Robert Schulz is supported by the Ministry of Economic Affairs, Labor and Tourism Baden-Wuerttemberg within the framework of the LaMaP project (No. WM33-42-47/145/4) and the LEAD-Loop project (No. BW1_4177/02). The work of Yuan Tian is supported by the National Key Research and Development Program of China (No. 2024YFC3210703).

**Funding** Open Access funding enabled and organized by Projekt DEAL.

**Data availability** Data will be made available on request.

## Declarations

**Conflict of interest** The authors declare that they have no known competing financial interests or personal relationships that could have appeared to influence the work reported in this paper.